\documentclass[review]{elsarticle}

\usepackage{geometry}
\journal{Journal of Information Sciences}

\usepackage{times}
\usepackage{soul}
\usepackage{url}
\usepackage[utf8]{inputenc}
\usepackage{bbding}
\usepackage{graphicx}
\usepackage{amsmath}
\usepackage{booktabs}
\usepackage{pifont}
\usepackage{graphicx}
\usepackage{subfigure}
\usepackage{multirow}
\usepackage{amsmath}
\usepackage{amsfonts}
\usepackage{color}
\usepackage{hyperref}
\usepackage{framed}

\usepackage[ruled, vlined, linesnumbered]{algorithm2e}

\graphicspath{ {images/} }

\begin{document}

\begin{frontmatter}

\title{EGC$^2$: Enhanced Graph Classification with Easy Graph Compression}

\author[mymainaddress,mysecondaryaddress]{Jinyin Chen}
\ead{chenjinyin@zjut.edu.cn}

\author[mysecondaryaddress]{Haiyang Xiong}
\ead{xhy19982021@163.com}

\author[mymainaddress]{Haibin Zheng}
\ead{haibinzheng320@gmail.com}

\author[mysecondaryaddress]{Dunjie Zhang}
\ead{zdj243166652@gmail.com}

\author[mythirdaryaddress]{Jian Zhang}
\ead{zj\_1994@outlook.com}

\author[myfourthaddress]{Mingwei Jia}
\ead{2112002168@zjut.edu.cn}

\author[myfourthaddress]{Yi Liu
\corref{mycorrespondingauthor}}
\cortext[mycorrespondingauthor]{Corresponding author}
\ead{yliuzju@zjut.edu.cn}

\address[mymainaddress]{Institute of Cyberspace Security, Zhejiang University of Technology, Hangzhou, 310023, China}
\address[mysecondaryaddress]{College of Information Engineering, Zhejiang University of Technology, Hangzhou, 310023, China}
\address[mythirdaryaddress]{School of Cyberspace Security, Hangzhou Dianzi University, Hangzhou, 310000, China}
\address[myfourthaddress]{Institute of Process Equipment and Control Engineering, Zhejiang University of Technology, Hangzhou, 310023, China}

\begin{abstract}
{\color{black}Graph classification is crucial in network analyses.}
{\color{black}Networks face potential security threats, such as adversarial attacks.
Some defense methods may trade off the algorithm complexity for robustness, such as adversarial training,
whereas others may trade off clean example performance, such as smoothing-based defense.
Most suffer from high complexity or low transferability.}
To address this problem,
we proposed EGC$^2$,
an enhanced graph classification model with easy graph compression.
{\color{black}EGC$^2$ captures the relationship between the features of different nodes by constructing feature graphs and improving the aggregation of the node-level representations.
To achieve lower-complexity defense applied to graph classification models,
EGC$^2$ utilizes a centrality-based edge-importance index to compress the graphs,
filtering out trivial structures and adversarial perturbations in the input graphs,
thus improving the model’s robustness.
Experiments on ten benchmark datasets demonstrate that the proposed feature read-out and graph compression mechanisms enhance the robustness of multiple basic models,
resulting in a state-of-the-art performance in terms of accuracy and robustness against various adversarial attacks.}

\end{abstract}


\end{frontmatter}

%\linenumbers

\section{Introduction}
{\color{black}Various graph-structured datasets,
representing complex relationships between objects such as social and biological networks~\cite{zhao2021community, doluca2021apal},
surround our lives.
Efficient graph mining methods are beneficial for making full use of these graphs.
Recently,
deep learning~\cite{zhang2019hierarchical} has been broadly applied in various fields,
such as computer vision~\cite{chen2020mag},
natural language processing~\cite{nguyen2022learning}, and graph learning~\cite{nguyen2022universal}.
The graph neural network (GNN)~\cite{kipf2017semi,peng2020spatial},
a prominent graph-mining tool, uses graph-structured data as input and learns the hidden representation of individual nodes or the entire graph. 
With wide applications,
GNN has demonstrated satisfactory learning abilities in different graph mining tasks,
such as node classification~\cite{zou2021embedding},
link predictions ~\cite{zhang2022knowledge},
and graph classification methods ~\cite{wang2021exploring}.}

{\color{black}To better aggregate the node-level representation learned by the GNNs at the graph level,
existing approaches sum up or average the node embeddings~\cite{kipf2017semi},
capture the information between the overall graphs and skip-gram model-based subgraph method~\cite{narayanan2017graph2vec},
or aggregate the node embedding by a deep learning architecture~\cite{zhang2019hierarchical}.
Furthermore,
to capture the hierarchical structure of the graph and aggregate the structural information of the graph layer by layer,
hierarchical graph-pooling methods~\cite{10.5555/3327345.3327389, lee2019self, zhang2019hierarchical} were designed to achieve state-of-the-art (SOTA) performance.
They outperform traditional graph kernel methods~\cite{1565664,10.5555/1953048.2078187},
or other graph-pooling methods~\cite{10.5555/3157382.3157527, zhang2018end} over most graph classification datasets.
Generally,
after pooling each graph,
they use mean-pooling or max-pooling as a read-out operation to aggregate the node-level representation into the graph level.
Although this process can efficiently generate graph level representation,
it ignores the different node contributions in the current pooling graph,
leading to the loss of structural information.

Since  numerous GNNs have been developed to achieve satisfactory performance in graph-related tasks, the robustness of GNNs \cite{jin2020certified,9322576} has captured wide attention. Both attacks and defenses of GNNs have been extensively studied.
In terms of attacks,
adversarial ~\cite{dai2018adversarial, chen2018fast, tang2020adversarial} and poisoning attacks~\cite{xi2020graph, zhang2020backdoor} have been proposed to explore the vulnerabilities of GNNs;
these have revealed the security issues of GNNs in inferencing and training processes,
respectively.
For defenses,
adversarial detection~\cite{10.1145/3394486.3403135} and robustness enhancement methods~\cite{tang2020adversarial} have been designed to address the threats of potential attacks.
Considering the attack scenario,
existing methods enhance the defense ability of graph classification models by introducing defense mechanisms, such as adversarial training~\cite{tang2020adversarial,jin2020certified} or random smoothing~\cite{9322576}.
These  defense mechanisms can improve robustness by modifying model parameters or structures.
When the graph classification model changes,
it necessitates redesigning the model structure or conducting adversarial training, which may increase defense costs.

In summary,
it remains a challenge to develop an end-to-end graph classification model that considers both classification performance and defense ability.
In this study,
we propose a robust graph classifier with defensibility
that achieves SOTA performance in graph classification and defensibility against adversarial attacks.
For graph classification,
graph-pooling methods have already learned different hierarchical pooling graphs.
Capturing the relationships between different node features can aggregate the graph structure information in these pooling graphs,
which leads to better performance.
Meanwhile,
designing a low-complexity and robust method for graph classification will provide secure applications.
What if there is an appreciable distinction between the actual contributed subgraph structures and adversarial perturbations?
We indeed observe this distinction in some edge importance indexes through the defined edge contribution importance (\emph{ECI}),
which inspired us to propose an effective and transferable defense with lower complexity,
while preserving the actual contributed subgraph structures.}
{\color{black}Consequently,
we propose EGC$^2$,
an enhanced graph classification model with easy graph compression.}
The main contributions of our study are summarized as follows:
\begin{itemize}
    \item {\color{black}We propose an enhanced graph classification model named EGC$^2$ which consists of the feature read-out and graph compression mechanisms.}
    EGC$^2$ considers both the graph classification performance and its robustness against adversarial attacks,
    thus achieving efficient and robust graph classification.
    \item {\color{black}By observing the distinction between the actual contributed subgraph structures and adversarial perturbations on the graph through the defined \emph{ECI},
    we propose a lightweight graph compression mechanism.
    Benefiting from the graph compression,
    we can improve the training speed, enhance the robustness of the model,
    and easily implement it because it is independent of model parameters.}
    \item To improve the performance of graph classification methods,
    we propose a feature read-out mechanism that aggregates high-order feature information of graphs.
    {\color{black}Furthermore,
    this mechanism has excellent transferability and can effectively improve the performance of different graph classification models.
}
\end{itemize}

\section{Related Works\label{RWs}}
{\color{black}Our work builds on two recent research categories.
Graph classification and its robustness.}
\subsection{Graph Classification}
{\color{black}In this section,
we briefly review graph classifiers,
mainly categorized into graph kernel,
GNN-based,
and graph pooling methods.}

%\textbf{\emph{Discriminative subgraph methods.}} In terms of the discriminative subgraph, Jin et al.~\cite{Ning10Ah} employed a novel subgraph encoding approach to support an arbitrary subgraph pattern exploration order and explores the subgraph pattern space in a process resembling biological evolution. To reduce the labeling cost for graph data, Kong et al.~\cite{Xian11Dual} maximized the dependency between subgraph features and graph labels using an active learning framework. Furthermore, Zhao et al.~\cite{Yu11Po} proposed an integrated approach to concurrently select the discriminative features and the negative graphs in an iterative manner. A new diversified discriminative score introduced by Zhu et al.~\cite{Yu12Graph} to select features that have a higher diversity.

\textbf{\emph{Graph kernel methods.}} {\color{black}Graph kernel methods~\cite{10.1145/2783258.2783417, borgwardt2020graph} assume that molecules with similar structures have similar functions and convert the core problem of graph learning to measure the similarity of different graphs.
Based on the shortest path in the graph,
Borgwardt et al.~\cite{1565664} defined the shortest-path (SP) kernel,
which is easy to calculate,
it is positive definite,
and suitable for various graphs.
To prevent discarding valuable information,
the Weisfeiler-Lehman (WL) kernel~\cite{10.5555/1953048.2078187} extracts features through the WL test of isomorphism on graphs.
For persistence summaries,
Zhao et al.~\cite{zhao2019learning} developed a weighted kernel method,
that is, WKPI,
and an optimization framework to learn the weight, which is positive semi-definite.}

\textbf{\emph{GNN-based methods.}}{\color{black}To learn an effective node-level representation
methods based on GNNs have also been proposed,
such as the graph convolutional network (GCN)~\cite{kipf2017semi} and graph attention network (GAT)~\cite{velivckovic2018graph}.
They obtained graph-level embeddings through maximum pooling or average pooling to achieve graph-classification tasks.
Furthermore,
Ullah et al.~\cite{ullah2022graph} introduced two strategies to enhance existing GNN frameworks,
that is, topological information enrichment through clustering coefficients and structural redesign of the network through the addition of dense layers.
UGformer, proposed by Nguyen et al.~\cite{nguyen2022universal} leverages the transformer on a set of sampled neighbors for each input node or transformer on all input nodes to learn precise graph representations.}

\textbf{\emph{Graph pooling methods.}} {\color{black}With the excellent performance of GNNs in the representation learning of graph-structured data,
graph-pooling methods have become popular in graph classification tasks.}
Graph topology-based pooling methods~\cite{10.5555/3157382.3157527, rhee2017hybrid} use the Graclus method as the pooling operation to calculate the clustering version for a given graph without graph features.
{\color{black}Considering the effect of graph features on the graph classification results,
global pooling methods usually pool node-level representations using summation or neural networks.}
Deep graph convolutional neural network (DGCNN)~\cite{zhang2018end} sorts the graph vertices in descending order,
then uses the sorted embedding to train traditional neural networks.
{\color{black}Hierarchical graph pooling methods learn hierarchical graph representations
to capture the graph structure information.
Ying et al.~\cite{10.5555/3327345.3327389} proposed a differentiable graph-pooling method called Diffpool
that aggregates the nodes into a new cluster as the input of the next layer using the cluster assignment matrix.
Because of the high complexity of the learning cluster assignment matrix,
the self-attention graph pooling (SAGPool)~\cite{lee2019self} considers both node features and graph topology.
The SAGPool selects the top-K nodes based on a self-attention mechanism to form the induced subgraph for the next input layer.
Hierarchical graph pooling with structure learning (HGP-SL) proposed by Zhang et al.~\cite{zhang2019hierarchical} adaptively developed the induced subgraphs through a nonparametric pooling operation,
and introduced a learning mechanism to preserve the integrity of the graph structure.

In general,
hierarchical graph pooling methods achieve better classification performance than other methods in most cases.
In addition to the well-performed hierarchical pooling operation,
they usually require a read-out operation to aggregate the features of the pooling graphs,
which also plays an essential role in the classifier, although it has rarely been studied.}

\subsection{Robustness of Graph Classification}
{\color{black}As attacks on GNNs are inevitable,
research on the robustness of GNNs in various graph-mining tasks has attracted considerable attention.
Here,
we briefly review both attacks and defenses.}

\textbf{\emph{Attacks on graph classification.}} {\color{black}There are two main types of attacks in graph classification: the adversarial attacks ~\cite{20Perturbations,20Restricted} in the inference process,
and poisoning during the training process.
During the inference process,
carefully crafted adversarial examples fool a graph classifier to output incorrect predictions.
Dai et al.~\cite{dai2018adversarial} proposed a reinforcement learning-based attack approach
and validated that GNNs are vulnerable to such attacks in graph classification tasks by learning a generalized attack policy.
Tang et al.~\cite{tang2020adversarial} used the pooling operation of hierarchical graph pooling neural networks (HGNNs) as the attack target.
They proposed a graph-pooling attack (PoolAttack),
thus deceiving HGNNs to retain incorrect node information by adding a slight perturbation to the original graph.}
The fast gradient attack (FGA) proposed by Chen et al.~\cite{chen2018fast} can also achieve efficient adversarial attacks based on the gradient information of target graph classification models.
{\color{black}To measure the robustness of certain graph-embedding methods,
Giordano et al.~\cite{giordano2022adversarial} manipulated the training data to produce an altered model through adversarial perturbations.
For the training process,
poisoning examples mixed with normal ones as training data were adopted to train the GNNs.
When the triggered example is input into the poisoning model,
it will also produce incorrect results.}
For example, works~\cite{zhang2020backdoor,xi2020graph, 21Explain_backdoor} used different methods to generate subgraphs as backdoor triggers and realized poisoning attacks on the graph classification model.

\textbf{\emph{Defenses on graph classification.}}
{\color{black}Because studies on attacks on graph classifiers have shown the vulnerability of the models,
other research aims to improve their robustness. 
Tang et al.~\cite{tang2020adversarial} conducted adversarial training on target HGNNs by generating adversarial examples,
thus improving the robustness of the HGNNs.
Gao et al.~\cite{9322576} proposed a certifiable robust method that smooths the decision boundary of the classic graph convolutional network (GCN) model by randomly perturbing graph-structured data and integrating predictions.
Jin et al.~\cite{jin2020certified} proposed a robustness certificate for GCN-based graph classification under topological perturbations,
which investigates the robustness of graph classification based on the Lagrangian duality and convex envelope.
These methods effectively improve the robustness of the graph classification model only if the quality and quantity of adversarial examples are promising~\cite{tang2020adversarial,jin2020certified},
or combined with a specific graph classification model~\cite{9322576}.
In general,
these defense methods aim to purposefully modify the model parameters,
thus enhancing the defense ability against adversarial attacks.
Although they satisfy the defense ability,
they also suffer from low transferability.}

\section{PRELIMINARIES}
This section introduces the problem definition of the graph classification,
the adversarial attack/defense on it and hierarchical graph pooling.
For convenience,
the definitions of some important symbols used are listed in TABLE \ref{tab:symbols data}.
\vspace{0.2cm}
\begin{table}[htb]
\renewcommand\arraystretch{0.8}
\small
	\centering
	\caption{THE DEFINITIONS OF SYMBOLS.}
	\label{tab:symbols data}
	\begin{tabular}{r|l}
		\toprule  \hline
		Notation        &Definition\\ \hline 
		 $A, A^{'}$ &  adjacency matrix / perturbation adjacency matrix\\
 		 $A_{pool}^{l+1}$ &   the ($l$+1)-th pooling adjacency matrix\\
		 $C^{l}$&   the $l$-th cluster assignment matrix \\
 		 $f_{\theta}(\cdot)$ &  graph classification model with parameters $\theta$\\
 		 $H_{pool}^{l+1}$&  the ($l$+1)-th pooling node-level matrix\\	 
		 $G=(V, E)$&   input graph $G$ with nodes $V$ and edges $E$\\
 		 $G$, ${G}^{com}$, ${G}^{'}$ &   original / compression / adversarial graph \\
		 $\mathcal{G}$, ${G}_{train}$ &      graph classification dataset, the training graphs\\
 		 $g$, $\hat{g}$ &   gradient matrix, symmetry gradient matrix\\
		 $Q$ &   category set of the graphs\\
		 $R_{g}$, $R_{CC}$ &   gradient vector, closeness centrality vector\\
		 $X, X^{'}$ &   node feature matrix / perturbation node feature matrix\\
 		 $y_{i}$ &      the $i$-th graph corresponding true label\\
		 $\sigma(\cdot)$ &  activation function\\
 		 $\Psi(G)$&      set of adversarial graphs for the original graph $G$\\
		\hline\bottomrule
	\end{tabular}
\end{table}

\subsection{Problem Definition}
 {\color{black}We represent a graph as $G=(V, E)$,
 where $V$ is the node set and
 $E$ is the edge set.}
 $G$ usually contains an attribute vector of each vertex.
 Here,
 we denote the attributes of the graph $G$ as $X$ and $A\in \{0,1\}^{n\times n}$ as the adjacency matrix,
 and use $G = (A, X)$ to represent a graph more concisely.
 
\textbf{DEFINITION 1 (Graph classification).}
A graph classification dataset $G_{set}$,
including $N$ graphs $\{G_1, G_2,..., G_N\}$.
The graph classification task aims to predict the categories of unlabeled graph through the model $f_{\theta}(\cdot)$  trained by the labeled graphs with its corresponding label.
 
{\color{black}The existing attack methods mainly occur in the inference process~\cite{dai2018adversarial, tang2020adversarial,20Perturbations,20Restricted},
which make the model predictions wrong through carefully designed tiny perturbations, i.e. adversarial attacks.
Therefore,
we focus on defending against attacks in the inference process.}

\textbf{DEFINITION 2 (Adversarial attack on graph classification).}
For a given target graph set $\mathcal{G}$ and the training graphs $G_{train}$,
$G_{t}^{'} = (A_{t}^{'},X_{t}^{'})$ is the adversarial graph generated by slightly perturbing the structure $A_{t}$ or attributes $X_{t}$ of the original target graph $G_{t}=(A_{t},X_{t})\in G_{set}$,
where $G_{train} \subset \mathcal{G}$.
The adversarial attack aims to maximize the loss of the target adversarial graph $G_{t}^{'}$ on $f_{\theta}(\cdot)$,
causing $G_{t}^{'}$ to get the wrong prediction result.
Here,
we focus on the adversarial attack on the graph structure,
which can be defined as:
\setlength{\parskip}{0\baselineskip}
 \begin{equation}\label{1}
\begin{array}{l}
\operatorname{maximize} \sum_{G^{\prime}_{t} \in \Psi(G)} \mathcal{L}\left(f_{\theta^{*}}\left(A^{\prime}_{t}, X_{t}\right), y_{t}\right) \\
\text { s.t. } \theta^{*}=\underset{\theta}{\arg \min } \sum_{G_{i} \in {G}_{train}} \mathcal{L}\left(f_{\theta}\left(A_{i}, X_{i}\right), y_{i}\right)
\end{array}
 \end{equation}
where $\Psi(G)$ is the set of adversarial graphs.
$f_{\theta}(\cdot)$ denotes the target graph classification model.

\textbf{DEFINITION 3 (Defense on graph classification).} For a given  training graphs $G_{train}$ and a graph classification model $f_{\theta}$,
the defender aims to minimize the loss of the adversarial graph $G_{t}^{'}\in \Psi(G)$ on $f_{\theta}(\cdot)$,
so that $G_{t}^{'}$ can get a correct prediction result.
To against various attacks,
the defense on graph classification can be defined as: \setlength{\parskip}{0\baselineskip}
 \begin{equation}\label{defense1}
\begin{array}{l}
\operatorname{minimize} \sum_{G^{\prime}_{t} \in \Psi(G)} \mathcal{L}\left(f_{\theta^{*}}\left(A^{\prime}_{t}, X_{t}\right), y_{t}\right) \\
\text { s.t. } \theta^{*}=\underset{\theta}{\arg \min } \sum_{G_{i} \in {G}_{train}} \mathcal{L}\left(f_{\theta}\left(A_{i}, X_{i}\right), y_{i}\right)
\end{array}
 \end{equation}
where ${G}_{train}$ is the set of a training graphs,
including the original graphs or the adversarial graphs.
$f_{\theta}$ is trained by ${G}_{train}$,
which can be a graph classification model with or without defense mechanism.
Here,
we consider ${G}_{train}$ as the original graphs and the $f_{\theta}$ as the graph classification model with a defense mechanism.

%\subsection{Basic Model of EGC\texorpdfstring{$^2$}{EGC\textasciicircum 2}: Diffpool}
\subsection{Diffpool}
As a graph classification method,
Diffpool~\cite{10.5555/3327345.3327389} introduces  a novel pooling operation for GNNs which can collect the  hierarchical information of graphs.
By using this pooling layer  with existing GNN models,
Diffpool achieves the SOTA performances on numerous real-world graph classification  datasets.
{\color{black}Therefore,
we use Diffpool as the  basic model of our approach,
but EGC$^2$ is not limited to it.}
Details are discussed in sections \ref{clean_example} and \ref{defense}. 

Diffpool employs the propagation function to implement convolution operations,
which can be expressed as:
\begin{equation}
	\label{eq2}
	H^{l,k}=\sigma({\hat{A}^l} H^{l,k-1} W^{l,k})
\end{equation}
\noindent where $\hat{A}^l=\tilde{D}^{-\frac{1}{2}} \tilde{A}^l \tilde{D}^{-\frac{1}{2}}$,
$A^l$ is the $l$-th adjacency matrix,
and $\tilde{A^l}=A^l+I^l_{N}$ is the adjacency matrix of the $l$-th pooling graph $G^l$ with self-connections.
{\color{black}$I^l_{N}$ is the identity matrix.
$\tilde{D}_{i i}=\sum_{j} \tilde{A^l}_{i j}$  denotes the degree matrix of $\tilde{A^l}$.}
$W^{l,k}$ is the parameters of the $k$-th propagation function in the $l$-th architecture.
$\sigma$ is the Relu active function.
The convolution operation of the  $l$-th architecture is denoted as:
\begin{equation}
	\label{eq3}
%	Z^l=GNN_{cov}(A^{l},H^{l})
	 H^{l}=GNN_{cov}(A^{l-1},H^{l-1};\theta^{l})
\end{equation}
where $H^{l}$ denotes the $l$-th node-level representation computed from $H^{l-1}$,
$\theta^{l}$ is the parameters set of the $l$-th architecture.
The input $A^{0}$ and $H^{0}$ are the original adjacency matrix and node attributes on the graph,
i.e., $A^{0}=A$, $H^{0}=X$,
respectively.

For the pooling operation,
the key idea is to learn the cluster assignment matrix over the nodes by GNNs.
{\color{black}The cluster assignment matrix of the $l$-th architecture is:
\begin{equation}
	\label{eq4}
	C^l={\rm softmax} \left ( GNN_{pool}(A^l,H^l) \right )
\end{equation}
where $GNN_{pool}$ has the same structure as $GNN_{cov}$.}
$C^{l} \in \mathbb{R}^{n_{l} \times n_{l+1}}$ denotes the cluster assignment matrix of $l$-th architecture and $n_{l+1}=n_{l} \times r$.
With the increase of assignment ratio $r$,
a larger pooling graph will be generated.

According to the adjacency matrix $A^l$,
the node-level representation $H^l$,
and the cluster assignment matrix $C^{l}$ in the $l$-th layer,
we generate a new pooling adjacency matrix $A_{pool}^{l+1}$ and a new pooling node-level representation $H_{pool}^{l+1}$ for the next layer.
Given these inputs,
the differentiable pooling process can be implemented according to the following two equations:
\begin{equation}
	\label{eq5}
	H_{pool}^{l+1}={C^{l}}^T H^l \in \mathbb{R}^{n_{l+1} \times d}
\end{equation}

\begin{equation}
	\label{eq6}
	A_{pool}^{l+1}={C^{l}}^T A^l C^{l} \in \mathbb{R}^{n_{l+1} \times n_{l+1}}
\end{equation}
where $d$ denotes the feature dimension of each node,
which is positively correlated to the training complexity of the model.
Eq.~\ref{eq5} aggregates the $H^l$ according to the cluster assignments $C^l$,
generating node-level representation for each of $n_{l+1}$ clusters.
Similarly,
Eq.~\ref{eq6} generates a pooling adjacency matrix based on the adjacency matrix $A^l$,
denoting the connectivity strength between each pair of clusters.

\begin{figure}
	\centering
	\includegraphics[width=1\linewidth]{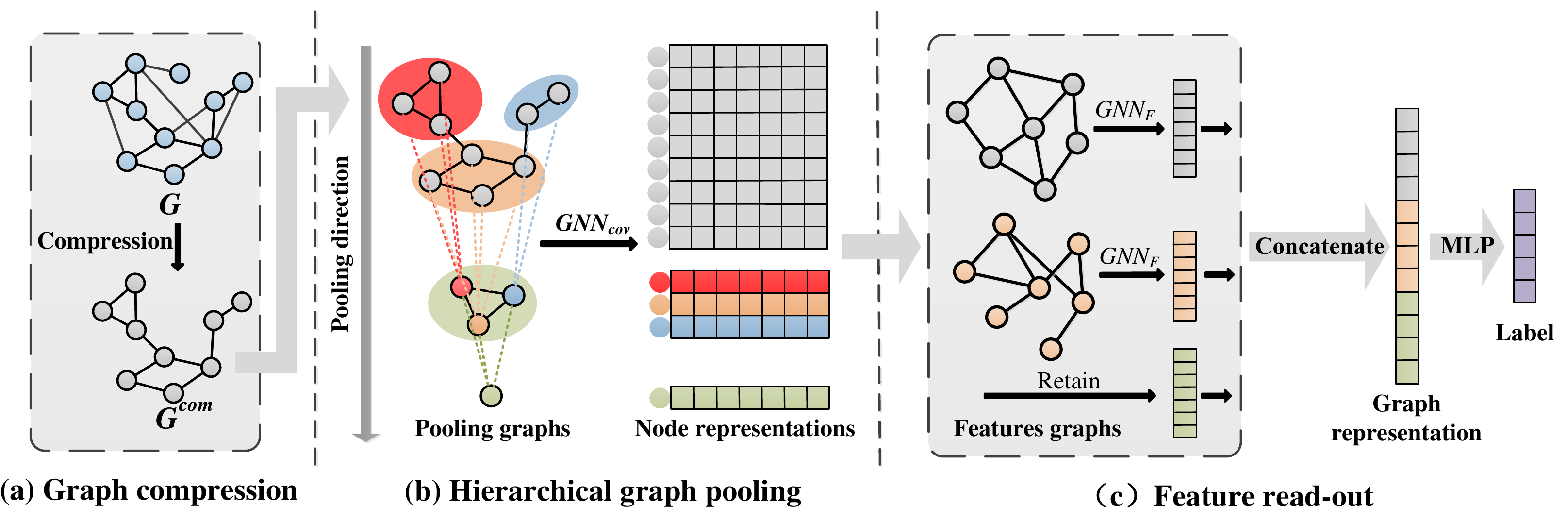}% 1\linewidth
	\caption{The main framework of EGC$^2$.
 {\color{black}It can be divided into three stages:
 (a) The graph compression preserves the actual contributed subgraph structures of $G$;
 (b) The hierarchical graph pooling learns the node-level representation of each pooling graph and constructs the corresponding feature graph;
 (c) The feature read-out aggregates node-level representations into graph-level,
 and outputs the graph label.}}\label{fig.1}
\end{figure} 

\section{Methodology}
\label{Met}
{\color{black}EGC$^2$ uses an end-to-end approach for graph classification, with robustness against adversarial attacks.
The specific architecture of EGC$^2$ is shown in Fig.\ref{fig.1}.
First,
EGC$^2$ compresses the original graph $G$ using the graph compression mechanism, which is used to improve the robustness of EGC$^2$ with scalability for a large graph,
denoted stage (a).
Subsequently, the feature read-out mechanism aggregates the node-level representation learned by the optional graph pooling operation into the graph level,
as shown in stage (b).
Finally, the final result of graph classification is obtained using a multilayer perceptron (MLP),
presented in stage (c).}

\subsection{Graph Compression Mechanism}
\label{4.3}
{\color{black}To introduce a lightweight and robust graph compression method for effective graph classification,
we evaluated the relationship between the contributions that the edges make to the model prediction and the importance calculated by the graph indexes.
Based on this relationship,
closely related graph indexes are used to extract critical edges and delete trivial edges from the graph that achieves compression,
as shown in Fig.\ref{fig.1}(a).

Specifically,
first,
the gradient method is used to quickly obtain the contribution value of the edges in the graph to the prediction result of the model,
to extract the actual contributed subgraph structure of the graph.
Second,
the graph indexes are used to obtain the importance value of the edges in the graph to extract the important edges.
Finally,
we analyzed the actual contributed subgraph structure and important edges using the defined \emph{ECI},
and proposed a lightweight graph compression method.}

\textbf{Analysis of Edge Contribution.}
\label{edge_con}
As an essential element on the graph classification model,
gradient can reflect the evolving relationship between variables and prediction results.
{\color{black}Consequently,
we obtain the contribution coefficients of the edges according to the gradient information of the end-to-end model,
and they indicate the contribution of each edge to the graph classification result.}
For a set of graphs $G_{set}$,
we use the cross-entropy function $\mathcal{L}$ to train EGC$^2$,
which can be represented as:
\begin{equation}\label{eq12}
	\mathcal{L} = -\sum_{G_i \in G_{set}}^{|G_{set}|} \sum_{j=1}^{|Q|}Y_{ij}\ln{\hat{Y}_{ij}\left(A_i,X_i\right)}
\end{equation}
where $Q=\left\{ \tau_{1},...,\tau_{|Q|}\right\}$ is the category set of the graphs,
$Y_{ij}$ is the  ground truth with $Y_{ij}=1$ if graph $G_i$ belongs to category $\tau_{j}$,
and $Y_{ij}=0$ otherwise.
$\hat{Y}_{ij}$ denotes the predicted probability that graph $G_i$ belongs to $\tau_{j}$.

{\color{black}According to Eq.~(\ref{eq12}),
the weight parameters of EGC$^2$ model are updated based on the gradient information,
so that the graph classification performance is steadily improved.}
Additionally,
the adjacency matrix $A_i$ and node attribute matrix $X_i$ are another group of variables in the loss function.
For graph classification task,
the adjacency matrix $A_i$ denotes the structural information of the graph $G_i$,
which can control the transmission of node attributes $X_i$.
In EGC$^2$,
$A_i$ plays a more critical role than $X_i$ in $\mathcal{L}$.
Thus,
we focus on using the gradient information to explore the relationship between $A_i$ and $\mathcal{L}$.

Based on the trained EGC$^2$ model,
we calculate the partial derivatives of the loss function $\mathcal{L}$ with respect to the adjacency matrix $A_{i}$,
and obtain the gradient matrix $g_i$,
represented by:
\begin{equation}
	\label{eq13}
	g_i=\frac{\partial \mathcal{L}_{i}}{\partial A_i}
\end{equation}

\begin{figure}
	\centering
	\includegraphics[width=0.55\linewidth]{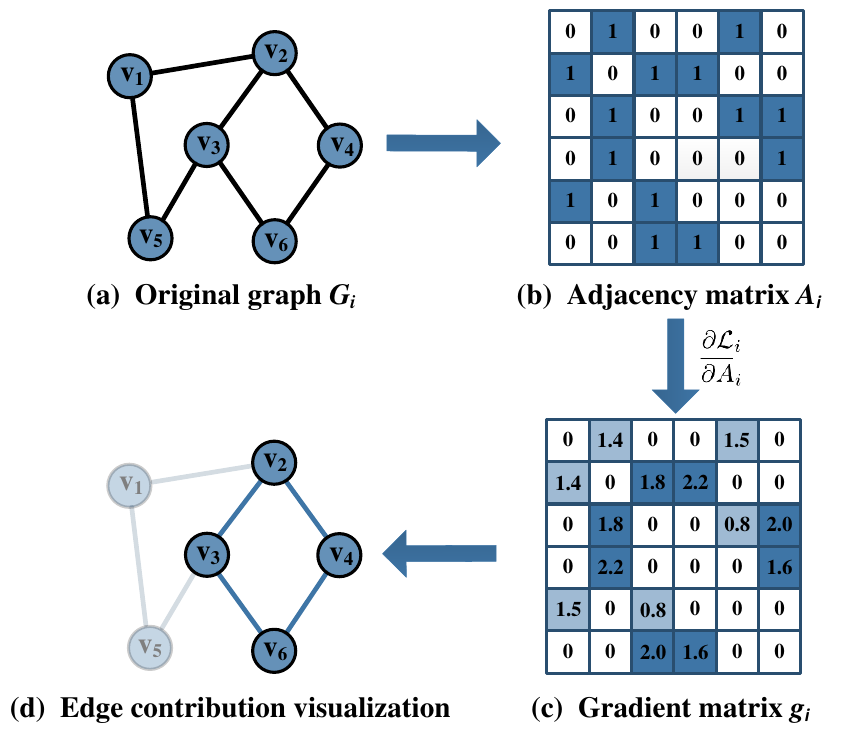}\\
	\caption{The processing of edge contribution visualization according to gradient information.}
	\label{fig.2}
\end{figure}

{\color{black}In Eq.~(\ref{eq13}),
each element $g_{i,jk}$ of $g_i$ corresponds to $A_{i,jk}$ of $A_i$.}
Generally,
the gradient value of $g_{i,jk}$ can reflect its contribution to the EGC$^2$.
Therefore,
the relationship between the classification results and the adjacency matrix $A_i$ can also be described according to the gradient matrix $g_i$.
Considering that the adjacent matrix of an undirected graph is symmetry,
we only focus on the relationship between $\mathcal{L}$ and the existing edges.
Here we adjust $g_i$ to obtain a symmetry gradient matrix $\hat{g}_i$.
\begin{equation}
	\label{eq14}
	\hat{g}_{i,jk}=\hat{g}_{i,kj}=\frac{g_{i,jk}+g_{i,kj}}{2}\cdot A_{i,jk}
\end{equation}
where $A_{i,jk}=1$ if node $j$ are connected with node $k$ and $A_{i,jk}=0$ otherwise.
$\hat{g}_{i,jk}$ consists of two parts: sign and magnitude.
Its positive (or negative) gradient value indicates that this edge has a positive (or negative) contribution for the graph classification.
Moreover,
the larger magnitude of $\hat{g}_{i,jk}$ means this edge has grater contribution for classification results.
{\color{black}According to Eq.~(\ref{eq14}),
we can use the non-zero values in the matrix $\hat{g}_i$ to intuitively evaluate the relationship between the graph classification results and the specific edges.}
Fig.~\ref{fig.2} shows the edge contribution according to gradient information.

{\color{black}To represent the contribution of each edge on the graph,
we transform $\hat{g}_{i}$ to a gradient vector $R_g$.}
The operation can be represented as:
\begin{equation}
	\label{eq15}
	R_g=\left (\hat{g}_{e_1}, \hat{g}_{e_2},...,\hat{g}_{e_j},...,\hat{g}_{e_{n_E}} \right )\in \mathbb{R}^{|E|}
\end{equation}
where $|E|$ denotes the number of edges.
The value of $j$-th element in $R_g$ is the absolute gradient value of the $j$-th edge on the graph.
$R_g$ takes edges as objects,
which measures the structural contribution of the entire graph.

\textbf{Analysis of Edge Importance.}
{\color{black}In addition to the edge contribution captured based on the model feedback,
some methods of the graph learning can also achieve edge importance analysis,
based on the local structure of the original graph instead of the gradient information.
They use the connection status of the graph to calculate the edge importance indexes,
such as degree centrality (DC)~\cite{camur2022star},
clustering coefficient (C)~\cite{bartesaghi2022tensor},
betweenness centrality (BC)~\cite{inoha2022efficient},
closeness centrality (CC)~\cite{zhao2017efficient} and  eigenvector centrality (EC)~\cite{roddenberry2021blind},
which are commonly used in various graph analyses.}
The detailed definition of these indexes are introduced in section \ref{edge_index}.
Therefore,
we can utilize these indexes and $R_g$ to explore the distinction between actual contributed subgraph structures and adversarial perturbations.

\textbf{Analysis of \emph{ECI}.}
In section \ref{edge_con},
we have analyzed the contribution of different edges to the result of graph classification.
In the research on the vulnerability of graph classification models,
the adversarial attack methods~\cite{tang2020adversarial,chen2018fast} mainly generate adversarial perturbations by optimizing the objective function.
Intuitively,
these adversarial perturbations usually have large gradient values,
so that they can effectively mislead the graph classification models.
Therefore,
considering only the edge contribution is not enough to distinguish between the actual contributed subgraph structures and the adversarial perturbations.
In this part,
we design an \emph{ECI} to capture the adversarial perturbations,
and propose a graph compression mechanism based on \emph{ECI} to improve the robustness of the graph classification model.

To obtain rich edge information,
some works~\cite{20LGESQL, 19SpeALine} convert the original graph into a line graph,
i.e., the nodes and links of the original graph are regarded as links and nodes of the line graph,
respectively.
By transforming the original graph to the line graph,
they can better mine the structural information of the graph,
thereby improving the performance of graph related tasks,
e.g., link prediction, network embedding.
Consequently,
for the graph $G_i=(V_i, E_i)$,
we first convert it into a line graph $L(G_i)=(E_i,D_i)$.
Then we calculate the CC vector $R_{CC}$, BC vector $R_{BC}$, EC vector $R_{EC}$ and C vector $R_{C}$,
respectively.
They are all in the $\mathbb{R}^{|E|}$ space,
which is the same as $R_g$.
Finally,
we utilize the cosine distance to calculate the similarity between these indexes and $R_g$,
as our \emph{ECI}.
Taking $R_{CC}$ as an example,
the \emph{ECI} is defined as:
\begin{equation}
\label{eq16}
  \emph{ECI}_{-R_{CC}}=\frac{{R^T_{CC}} \cdot R_{g}}{\left \|R_{CC} \right \| \cdot \left \|R_{g} \right \|}
\end{equation}
where $\left \| \cdot \right \|$ denotes the length of the vector.
\emph{ECI}$_{-R_{CC}} \in [-1,1]$,
the larger value of \emph{ECI}$_{-R_{CC}}$ indicates that $R_{CC}$ and $R_{g}$ have higher correlation,
which means that $R_{CC}$ has a stronger ability to capture essential structural information on the original graph.
The analysis of different \emph{ECI} will be given in section~\ref{5.6}.

{\color{black}Suppose the edge importance index such as $R_{CC}$ is evaluated as highly correlated with $R_g$.
In that case,
we can conclude that the $R_{CC}$ fits the end-to-end model and can effectively extract the structure that plays a crucial role in graph classification.
To capture the possible adversarial perturbations from the edges with large contribution values in the adversarial examples,
we calculate the \emph{ECI} of the clean examples and the adversarial examples,
respectively,
and evaluate the difference between them by Eq.~(\ref{eq17}):}
\begin{equation}
\label{eq17}
  \triangle = \emph{ECI}^{ clean} - \emph{ECI}^{ adv}
\end{equation}
where \emph{ECI}$^{ clean}$ and \emph{ECI}$^{ adv}$ calculate the \emph{ECI} value of the clean examples and the adversarial examples under an important index,
respectively.
{\color{black}The intuition of Eq.~(\ref{eq17}) is that the actual contributed subgraph structure on a clean graph often has a large contribution value and edge importance at the same time.}
Due to the neglect of the original graph structure characteristics during the generation process,
the adversarial perturbations may have a smaller edge importance value,
although they are also in a considerable contribution value.
We discuss this speculate through experiments in section~\ref{5.6}.

\begin{figure}
  \centering
  \includegraphics[width=0.7\linewidth]{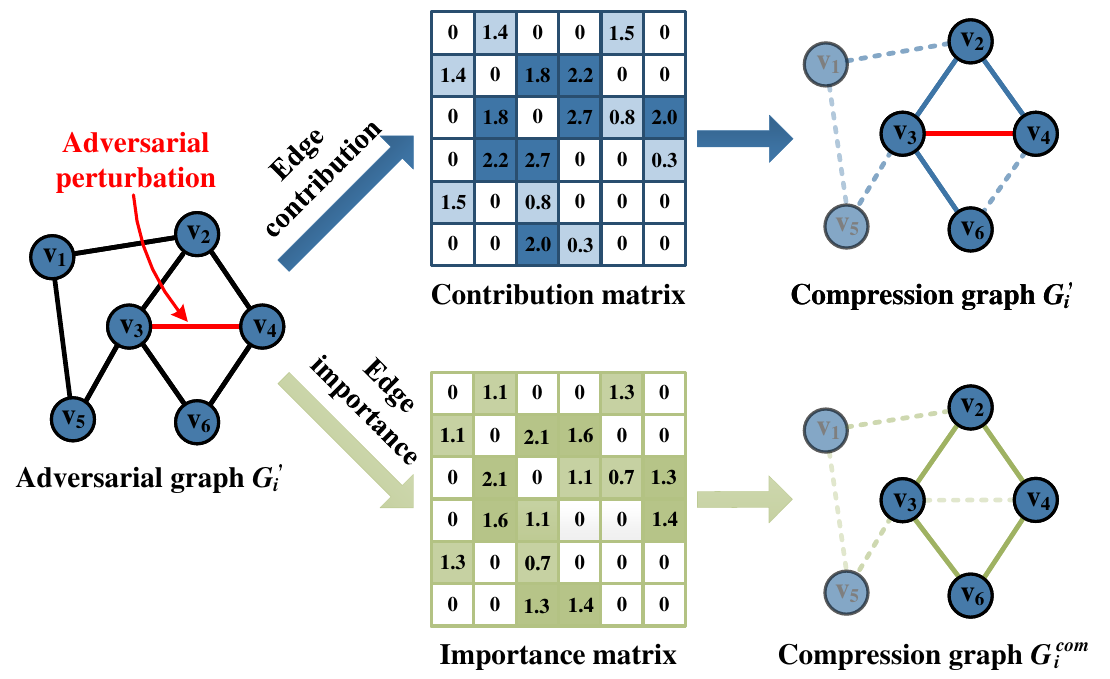}\\
  \caption{Graph compression by edge contribution or edge importance,
  respectively.
  Compared with the contribution-based graph compression,
  the importance-based graph compression can better preserve the actual contributed subgraph structures.}
  \label{fig.3}
\end{figure}

We calculate the \emph{ECI} and the $\triangle$ under different edge importance indexes,
and still take $R_ {CC} $ as an example.
If $R_{CC}$ has a large \emph{ECI} value in clean examples and a large $\triangle$ between clean examples and adversarial examples,
then we will choose it to guide graph compression.
Specifically,
for the adversarial example $G_{t}^{'}$,
we will sort the edges of $G_{t}^{'}$ according to $R_{CC}$,
and delete the lower-ranked edges and isolated nodes by the compression ratio $\gamma$.
We take $\gamma = 50\%$ as an example to show the process of graph compression in Fig.~\ref{fig.3}.
Combined with the edge contribution of the clean graph in Fig.~\ref{fig.2},
when selecting the edge contribution to guide the graph compression,
the adversarial perturbations on the input graph may be preserved instead of the actual essential edges.
On the contrary,
an appropriate edge importance index can preserve the actual essential edges and filter adversarial perturbations.
After the graph compression,
we can obtain a simpler graph classification dataset with the actual contributed subgraph structures,
which improves the training speed and robustness of the model.

\subsection{Hierarchical Graph Pooling}
%\textbf{Hierarchical graph pooling.}
Taking advantage of the transferability of the feature read-out mechanism,
EGC$^2$ can learn the node-level representation of the input graph with multiple basic models.
They can learn effective hierarchical graph structures and the corresponding node-level representations by a hierarchical graph pooling operation.
Here,
as shown in Fig.\ref{fig.1}(b),
we learn from but not limited to the idea of Diffpool~\cite{10.5555/3327345.3327389} to learn the hierarchical structure of the graph.

\subsection{Feature Read-out Mechanism}
%\subsection{Model Structure of EGC\texorpdfstring{$^2$}{EGC\textasciicircum 2}}
%\label{Feature Read-out}
Benefiting from the graph compression mechanism for robust and efficient graph classification,
the model may encounter the challenge of performance degradation.
Besides,
the existing read-out mechanisms~\cite{kipf2017semi,velivckovic2018graph,lee2019self} have the average,
sum,
maximum or minimum strategies to get the graph-level representation.
However,
they may lose important information in the graphs by simply deleting,
summing,
or averaging node information.
Inspired by the work~\cite{wang2020gcn},
they adapted the model to learn the correlation information between topology and node features by transforming feature graphs,
which effectively improved the performance of the model.

To address the issue,
we propose a feature read-out mechanism to enhance the ability of the model to extract graph features to achieve more efficient graph classification.
In Fig.\ref{fig.1}(c),
it preserves the graph features in the process of hierarchical pooling and reconstructs the feature graphs.
The graph neural network extracts the information from the feature graphs so that the model can adaptively learn the appropriate graph-level feature.

For the aforementioned considerations,
we construct hierarchical feature graphs $G_F^l=(A_F^l,X_F^l)$ based on node-level representations $H_{pool}^{l}$, $l=\{0,1,2\}$ to learn the relationships between different features.
We introduce a novel concept of \emph{feature nodes}.
Specifically,
we regard each dimension of the hidden features of the $l$-th node-level representation as $d$ \emph{feature nodes} $V_F^l$,
and consider the original nodes as the hidden features of the \emph{feature node} at the same time,
to obtain the node-level representation of the feature graph:
\begin{equation}
	\label{eq7}
	X_F^l= (H_{pool}^l)^T \in R^{d \times n_l}
\end{equation}

To obtain the topological structure of the feature graphs,
we calculate the similarity between different \emph{feature nodes},
and add edges to the \emph{feature nodes} pairs with a high similarity:
\begin{equation}
	\label{eq8}
	A_F^l=top(\frac{X_F^l \cdot (X_F^l)^T}{\left\|X_F^l\right\|\left\|(X_F^l)^T\right\|},k) \in R^{d \times d}
\end{equation}
where $\left \| \cdot \right \|$ is the length of the vector.
The feature edge ratio $k$ determines the scale of the feature graph,
which will be set according to the experimental results.
Here $top(S,k)$ means that the values in the matrix $S \in R^{d \times d}$ are sorted in descending order.
Select the top $k\times d \times d$ values as 1 and the other values as 0,
where $k \in(0,1]$.
Then utilizing the feature graphs as input,
the feature aggregation graph neural network $GNN_F$ aggregates them into a graph-level representation $Z_G^{l}\in \mathbb{R}^{d \times 1}$:
\begin{equation}
	\label{eq9}
	Z_G^{l}=GNN_F(A_F^{l},X_F^{l};\theta_F^{l})
\end{equation}
where $GNN_F$ has the same structure as $GNNcov$,
and $\theta_F^l$ is the parameters of the $l$-th feature read-out mechanism.

{\color{black}It is worth noting that although our feature read-out mechanism introduces additional parameters,
the increase in the complexity of the graph classification model is controllable,
since the feature dimension $d$ is a hyperparameter that can be easily selected based on experimental results.

After obtaining the graph-level representations for different levels of pooling graphs,
we concatenate them together into the global-level graph representation:}
\begin{equation}
	\label{eq10}
	Z_G=concat(Z_G^{1},Z_G^{2},...,Z_G^{L})
\end{equation}
where $concat(\cdot)$ denotes the concatenate function,
which stitches the graph-level representations of $L$ architecture.
{\color{black}Finally,
we take the global-level representation as the input of the MLP layer with a softmax classifier:}
\begin{equation}
	\label{eq11}
	\hat{Y}= softmax(MLP(Z_G))
\end{equation}
This final predicted probability can then be used to build a cross-entropy function with the actual label of the graphs.
The entire system is an end-to-end model that can be trained by stochastic gradient descent.

\subsection{EGC\texorpdfstring{$^2$}{EGC\textasciicircum 2} Algorithm}
{\color{black}In summary,
EGC$^2$ achieves the robust and high-performance graph classification through the following three stages.
First, select an appropriate edge importance index for graph compression based on \emph{ECI} (In the absence of adversarial examples,
$C$ can be directly used to guide graph compression according to the experimental results). 
Then obtain the compression graph $G^{com}$,
which reduces the training complexity and improves the robustness of the model.
Second, take the compression graph $G^{com}$ as the input of EGC$^2$,
the hierarchical graph pooling layer learns different pooling graphs and corresponding node-level representations.
At last, the feature read-out mechanism enhances the effect of aggregating node-level representations into the global-level,
thereby improving the performance of graph classification.
The pseudocode of EGC$^2$ is presented in Algorithm \ref{algorithm_1}.}

\begin{algorithm}[t]
	\caption{EGC2: Enhanced Graph Classification with Easy Graph Compression} \label{algorithm_1}
	\begin{small}
		\BlankLine
		\KwIn{Graph classification dataset $\mathcal{G}$, graph convolution model $GNN_{cov}$, graph pooling model $GNN_{pool}$, feature aggregation model $GNN_{F}$, graph assignment ratio $r$}
		\KwOut{The predicted probability list $Y_{all}$}
		Initialize the predicted probability empty list $Y_{all}$.
		
		\While{$G_i \in \mathcal{G}$}{
			Generate a compression graph $G_i^{c o m}$ by Eq. (\ref{eq16}) and Eq. (\ref{eq17}).
			
			Compute the $l$-th node-level representation $H^{l}$ and the $l$-th cluster assignment matrix $C^{l}$ by Eq. (\ref{eq3}) and Eq. (\ref{eq4}).
			
			Update $H_{\text {pool }}^{l+1}=C^{l^T} H^l$ and $A_{p o o l}^{l+1}=C^{l^T} A^l C^l$ by Eq. (\ref{eq5}) and Eq. (\ref{eq6}).
			
			Generate the $l$-th node-level representation of the feature graph $X_F^l$ by Eq. (\ref{eq7}).
			
			Generate the $l$-th adjacency matrix of the feature graph $A_F^l$ by Eq. (\ref{eq8}).
			
			Get the $l$-th graph-level representation of the feature graph $Z_G^l=G N N_F\left(A_F^l, X_F^l ; \theta_F^l\right)$ by Eq. (\ref{eq9}).
			
			Compute $Z_G=\operatorname{concat}\left(Z_G^1, Z_G^2, \ldots, Z_G^L\right)$ by Eq. (\ref{eq10}).
			
			Obtain the predicted probability $\hat{Y}=\operatorname{softmax}\left(M L P\left(Z_G\right)\right)$ by Eq. (\ref{eq11}).
			
			Add $\hat{Y}$ to $Y_{all}$.
		}
		Return $Y_{all}$.
	\end{small}
\end{algorithm}

\section{Experiments}
\label{exp}
To evaluate our proposed approach,
we conduct experiments to answer the following research questions (RQs):
\begin{itemize}
    \item \textbf{RQ1.} Can the proposed \emph{ECI} help capture the distinction between the actual contributed subgraph structures and the adversarial perturbations?
    \item \textbf{RQ2.} Does the proposed EGC$^2$ achieve the state-of-the-art performance of graph classification?
    \item \textbf{RQ3.} How does the defense performance of EGC$^2$  under adversarial attacks of various attack budgets?
    \item \textbf{RQ4.} Does EGC$^2$ have transferable defense for multiple graph classification methods or attack scenarios?
\end{itemize}

\subsection{Settings}
Following some previous works~\cite{10.5555/3327345.3327389, lee2019self, zhang2019hierarchical},
we evaluate the graph classification methods using 10-fold cross validation.
In our EGC$^2$,
we implement three hierarchical graph pooling layers,
and each hierarchical graph pooling layer is followed by a feature read-out mechanism.
The MLP consists of two fully connected layers with number of neurons in each layer setting as 64 and 2,
followed by a softmax classifier.
We use the Adam optimizer to optimize the model,
and the learning rate is searched in \{0.1,0.01,0.001\}.
Other hyper-parameters are set by a hyper-parameter search in the feature edge ratio $k \in [0.05,0.75]$,
the feature node number $d\in \{32,64,128,256,512\}$ (also used as the dimension of node representations),
and the assignment ratio $r \in [0.1,0.9]$.
We employ the early stopping criterion during the training process,
i.e., we stop training if the validation loss does not decrease for 100 consecutive epochs.
We appropriately oversample the examples with class label `1' in the data to alleviate the uneven distribution problem  of the OGBL-MOLHIV dataset.
We implement EGC$^2$ with PyTorch,
and our experimental environment consists of i7-7700K 3.5GHzx8 (CPU),
TITAN Xp 12GiB (GPU),
16GBx4 memory (DDR4) and Ubuntu 16.04 (OS).
The code and data of EGC$^2$ could be downloaded from https://github.com/Seaocn/EGC2.

\subsection{Datasets}
{\color{black}To test the performance of EGC$^2$,
we select ten real-world graph datasets to conduct experiments.
Seven datasets are about bio-informatics,
chemo-informatics and computer vision,
including PTC~\cite{nguyen2022universal},
PROTEINS~\cite{zhang2019hierarchical},
D\&D~\cite{10.5555/3327345.3327389},
NCI1~\cite{10.5555/1953048.2078187},
NCI109~\cite{10.5555/1953048.2078187},
OGBL-MOLHIV~\cite{20OGBL} and FIRSTMM\_DB~\cite{2013GraphKernels}.
The rest of them are social network datasets,
including IMDB-BINARY~\cite{10.1145/2783258.2783417},
REDDIT-BINARY~\cite{10.1145/2783258.2783417} and  TWITCH\_EGOS~\cite{An_API}.
The basic statistics are summarized in TABLE \ref{tab:1}.}

\begin{table}
	\centering
	\small
	\renewcommand\arraystretch{0.8}
	\caption{The basic statistics of ten datasets.}
	
		\begin{tabular}{c|ccccc}
			\hline \hline
			Datasets       & \#Graphs & \#Classes & \#Ave\_nodes    & \#Ave\_edges & \#Graphs{[}Class{]}    \\ \hline
			PTC           & 344      & 2         & 14.29             & 14.69       & 152{[}0{]}, 192{[}1{]}   \\ \hline
			PROTEINS      & 1113     & 2         & 39.06             & 72.82       & 663{[}0{]}, 450{[}1{]}   \\ \hline
			DD            & 1178     & 2         & 284.32            & 62.14       & 692{[}0{]}, 486{[}1{]}   \\ \hline
			NCI1          & 4110     & 2         & 29.87             & 32.30        & 2053{[}0{]}, 2057{[}1{]}    \\ \hline
			NCI109        & 4127     & 2         & 29.69             & 32.13        & 2045{[}0{]}, 2082{[}1{]}     \\ \hline
			IMDB-BINARY   & 1000     & 2         & 19.77             & 96.53          & 500{[}0{]}, 500{[}1{]}  \\ \hline
			REDDIT-BINARY & 2000     & 2         & 429.63            & 497.75        & 1000{[}0{]}, 1000{[}1{]}     \\ \hline
			 OGBL-MOLHIV &  41127     &  2         &  25.50           &   27.50       & 39598{[}0{]}, 1529{[}1{]}  \\ \hline
			 TWITCH\_EGOS &   127094     &   2         &   29.67            &   86.59     & 58761{[}0{]}, 68333{[}1{]}     \\ \hline
			 FIRSTMM\_DB &   41    &   11        &   1377.27	            &   3074.10      & 17{[}0-3{]}, 12{[}4-7{]}, 12{[}8-10{]}   \\ \hline
			\hline
		\end{tabular}
	
	\label{tab:1}
\end{table}

\subsection{Graph Classification Baselines}
To testify the classification performance of EGC$^2$,
we choose three groups of graph classification methods as  baselines,
including graph kernel methods, GNN-based methods, and graph pooling methods.

\begin{itemize}
    \item \textbf{Graph kernel methods.} This group of methods realize graph classification by constructing graph kernels that can measure the similarity of graphs.
    Our baselines contain three  graph kernel algorithms: SP~\cite{1565664},
    WL~\cite{10.5555/1953048.2078187} and WKPI~~\cite{zhao2019learning}.
    \item \textbf{GNN-based methods.} GNNs such as GCN~\cite{kipf2017semi} and GAT~\cite{velivckovic2018graph} are designed to learn the effective node-level representation.
    Furthermore,
    UGformer~\cite{nguyen2022universal} leverages the transformer on a set of sampled neighbors for each input node or the transformer on all input nodes.
    In this group of baselines,
    we obtain graph-level representation through mean-pooling,
    sum-pooling or max-pooling.
    \item \textbf{Graph pooling methods.} The graph pooling methods introduce the pooling operation into GNNs to better learn graph-level representation.
    We take the best-performing hierarchical graph pooling methods as the baselines,
    where Diffpool~\cite{10.5555/3327345.3327389} is based on node grouping,
    and SAGPool~\cite{lee2019self} and HGP-SL~\cite{zhang2019hierarchical} are based on node sampling.
\end{itemize}

\textbf{Baseline parameter settings.} In the graph kernel methods,
we compute the classification accuracy using the $C$-SVM implementation of LIBSVM~\cite{10.1145/1961189.1961199}.
The $C$ parameter was selected from $\{10^{-3}, 10^{-2},...,$\par \noindent$10^2, 10^3\}$.
Based on the source code released by the authors and their suggested parameter settings,
we use an additional pooling operation and the same MLP as in EGC$^2$ to obtain the final graph classification results for the GNN-based methods.
For all the graph pooling methods,
we use 10-fold cross validation results reported by the authors.

\subsection{Adversarial Attack Methods}
\label{sec5.4}
To test whether our \emph{ECI} can capture the adversarial perturbations,
and verify the robustness of our EGC$^2$,
we add adversarial perturbations with different perturbation ratios (from 5$\%$ to 40$\%$) of the existing edges to the test set.
The attack methods with different attack knowledge,
i.e. white-box attacks and black-box attacks,
are briefly described as follows:
\begin{itemize}
    \item \textbf{FGA~\cite{chen2018fast}.} As an efficient and convenient white-box attack method,
    FGA utilizes the gradient information of pairwise nodes based on the trained GNN model to generate adversarial examples to attack.
    \item \textbf{PoolAttack~\cite{tang2020adversarial}.} Due to the difficulty of obtaining the gradient information of the hierarchical graph pooling methods,
    PoolAttack as the black-box attack method iteratively generates adversarial perturbations to fool the pooling operation in hierarchical graph pooling models through the surrogate model,
    thus reserving the wrong nodes.
\end{itemize}

\subsection{Edge Attributes}
\label{edge_index}
The node centrality in the line graph can reflect the importance of edges,
including degree centrality,
clustering coefficient,
betweenness centrality,
closeness centrality and eigenvector centrality. {\color{black}They are defined and calculated as follows:
\begin{itemize}
    \item \textbf{Degree Centrality (DC)~\cite{camur2022star}.} DC is defined as:
    \begin{equation}
    	D C_{i}=\frac{d_{i}}{N-1}
    \end{equation}
    where $d_{i}$ is the degree of node $i$ and $N$ is the number of nodes in the line graph.
    
    \item \textbf{Clustering Coefficient (C)~\cite{bartesaghi2022tensor}.} $L_{i}$ is the number of links between the $k_{i}$ neighbors of node $i$. C can be defined as:
    \begin{equation}
    	C_{i}=\frac{2 L_{i}}{k_{i}\left(k_{i}-1\right)}
    \end{equation}
    
    \item \textbf{Betweenness Centrality (BC)~\cite{inoha2022efficient}.}
    It is defined as: 
    \begin{equation}
    	B C_{i}=\sum_{s \neq i \neq t} \frac{n_{s t}^{i}}{\sigma_{s t}}
    \end{equation}
    where $\sigma_{s t}$ is the total number of the shortest paths between nodes $s$ and $t$ in the line graph,
    and $n^{i}_{s t}$ represents the number of the shortest paths between nodes $s$ and $t$ that pass through the node $i$.
    
    \item \textbf{Closeness Centrality (CC)~\cite{zhao2017efficient}.} It is defined as:
    \begin{equation}
    	C C_{i}=\frac{N}{\sum_{j=1}^{N} g_{i j}}
    \end{equation}
    where $g_{i j}$ denotes the shortest path length between nodes $i$ and $j$ in the line graph.
    The shorter the distances between the node $i$ and the rest nodes are,
    the more central the node $i$ is.
    
    \item \textbf{Eigenvector Centrality (EC)~\cite{roddenberry2021blind}.} EC is defined as:
    \begin{equation}
    	E C_{i}=\alpha \sum_{j=1}^{N} a_{i j} E C_{j}
    \end{equation}
    where $a_{i j}$ is the element of the adjacency matrix of the line graph,
    i.e., $a_{i j}=1$ if nodes $i$ and $j$ are connected and $a_{i j}=0$ otherwise,
    and $\alpha$ is a non-zero constant.
\end{itemize}
}
\subsection{Evaluation Metrics}
To evaluate the graph classification performance and defense ability of EGC$^2$,
we use the following three metrics:
\begin{itemize}
    \item \textbf{Accuracy.~\cite{lee2019self}} As one commonly used evaluation metrics,
    accuracy measures the quality of results based on the percentage of correct predictions over total instances.
    It is defined as:
    \begin{equation}
    	\text { Accuracy }=\frac{N_{\text {cor }}}{N_{\text {total }}}
    \end{equation}
    where $N_{cor}$ and $N_total$ are the number of samples correctly predicted and the total number of samples predicted,
    by the graph classification methods,
    respectively.
    
    \item \textbf{Balanced Accuracy.~\cite{Is15De}} To deal the issue with imbalanced datasets,
    balanced accuracy is defined as the average recall obtained on each class:
    \begin{equation}
    	\text { Balanced accuracy }=\frac{R_{Sensitivity}+R_{Specificity}}{2}
    \end{equation}
    where $R_{Sensitivity}$ is the percentage of positive cases that the graph classification method is able to detect.
    $R_{Specificity}$ is the percentage of negative cases that the graph classification method is able to detect.

    \item \textbf{Attack success rate (ASR)~\cite{tang2020adversarial}.} It represents the rate of the targets that will be successfully attacked under a given constraint,
    which is defined as:
    \begin{small}
      \begin{equation}
\label{eq19}
  \rm {ASR=\frac{Number \ of \ successfully \ attacked \ graphs}{Number \ of \  attacked \ graphs}}
\end{equation}
\end{small}
\end{itemize}

\begin{figure}[htbp]
  \centering
  \includegraphics[width=0.9\linewidth]{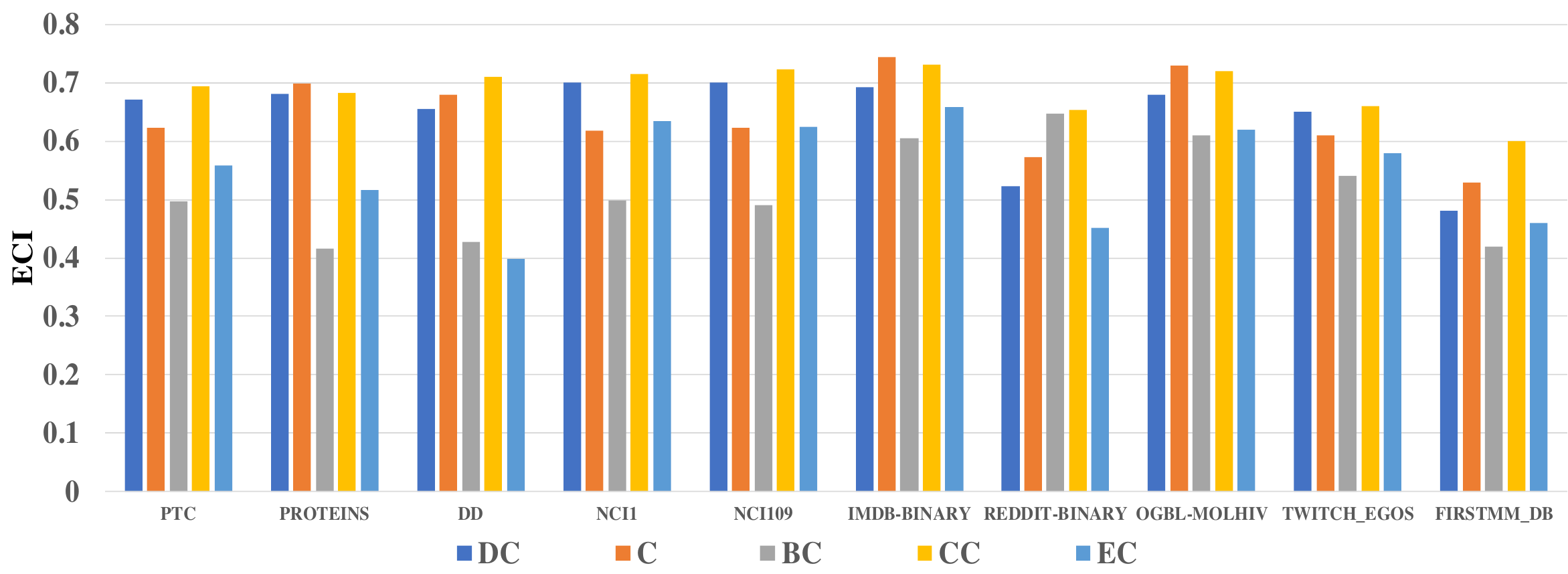}\\
  \caption{The different \emph{ECI} for each dataset.}
  \label{fig.5}
\end{figure}

\subsection{Edge Distinction Analysis}
\label{5.6}
{\color{black}In this subsection,
we focus on question \textbf{RQ1} and investigate the distinction between the actual contributed subgraph structures and adversarial perturbations through the proposed \emph{ECI}.
This provides a practical basis for using the graph compression mechanism to achieve effective and transferable defense with low complexity.}

\subsubsection{Correlation between edge contribution and importance indexes}
{\color{black}We first used Eq.(\ref{eq16}) to calculate the \emph{ECI} between the edge contribution and different edge importance indexes for each dataset.
The experimental results are presented in Fig. \ref{fig.5}.
For the ten datasets,
compared to the other two edge importance indexes:
DC, C, and CC are highly correlated with the edge contribution,
and the average \emph{ECI} values across the datasets are approximately 0.65 (DC), 0.64 (C), and 0.68 (CC).
This observation provides solid evidence that the important edges obtained by the edge importance indexes are consistent with those obtained by the edge contribution.
The edges that significantly contribute to the graph classification result typically have larger edge importance.
It also verifies that it is reasonable to preserve the critical structure of the graph through edge importance indexes.}

\begin{figure}[htbp]
  \centering
  \includegraphics[width=1\linewidth]{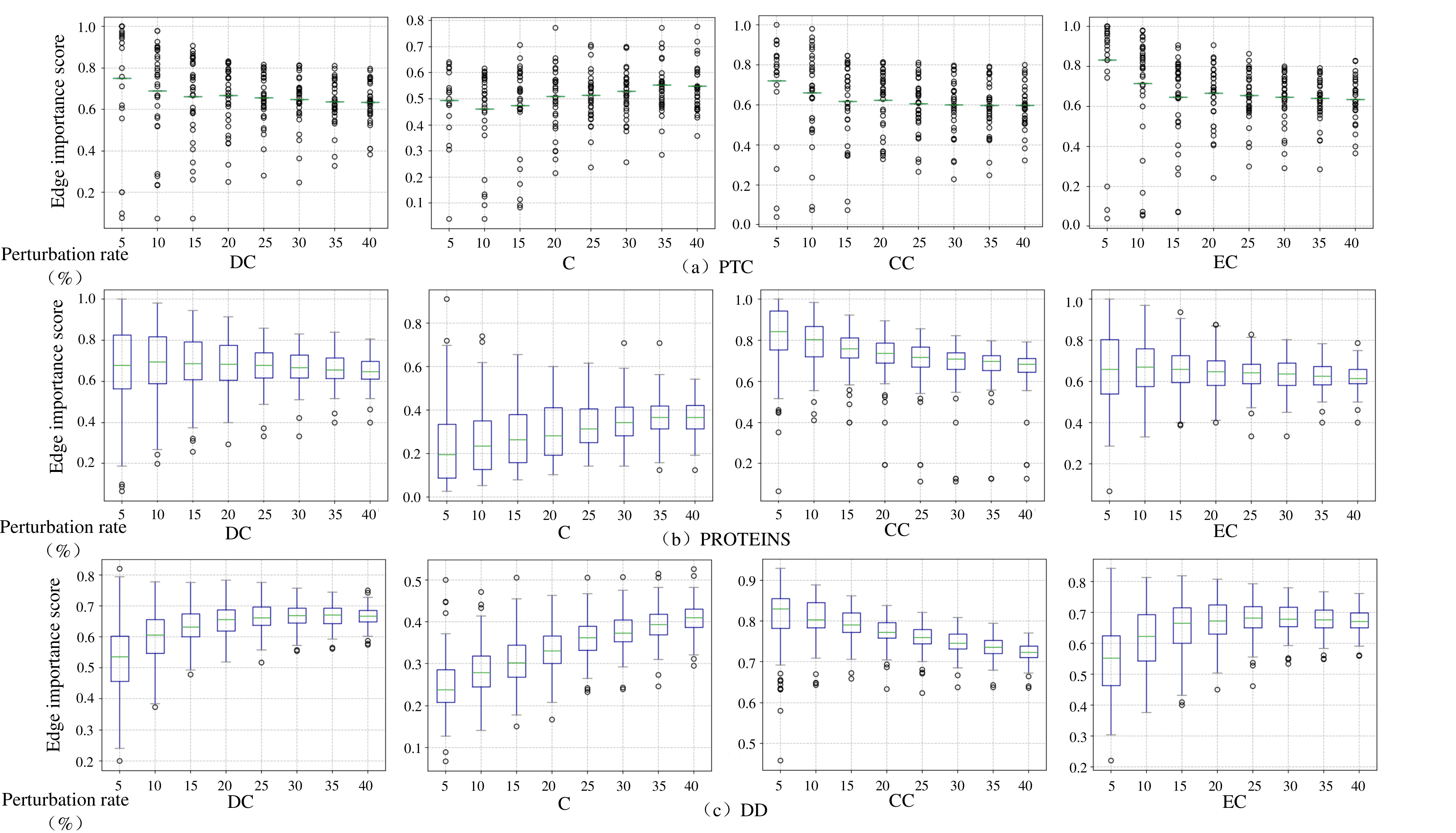}% 1\linewidth
  \caption{The edge importance of the adversarial perturbations generated by FGA.
  The x-axis represents the perturbation ratio.
  The y-axis represents the edge importance score of the adversarial perturbations,
  a larger score means the edge has greater edge importance.}
  \label{fig.6}
\end{figure}

\subsubsection{The edge importance of adversarial perturbations} 
{\color{black}As described in section \ref{4.3},
adversarial perturbations generated by FGA and PoolAttack typically have a large edge contribution
because they all use the gradient information of the objective function to achieve adversarial attacks.
Based on these considerations,
we select edge importance indexes (i.e., DC, C, CC, and EC) that are highly correlated with the edge contribution to exploring the distinction between adversarial perturbations and the actual contributed subgraph structures.

In Fig. \ref{fig.6},
we evaluated the adversarial perturbations generated by the FGA through different edge importance indexes.
We designed an edge importance score with a value of (0--1) according to the descending order of the edge importance indexes.
A larger score indicates a greater edge importance.
In the plot,
each dot represents the average edge importance score of adversarial perturbations in each graph.
Most adversarial perturbations also have high edge importance scores,
which further verifies the correlation between the edge contribution and edge importance indexes.
Adversarial perturbations have a small C value,
which is different from when the edge contribution is highly correlated with edge importance indexes.
Therefore,
we conclude that the actual contributed subgraph structures and adversarial perturbations on the graph often have an appreciable distinction in C.
Adversarial perturbations can be better captured through this distinction.
The same conclusion can also be reached in the remaining datasets and  adversarial examples generated by PoolAttack.}

\begin{framed}
{\color{black}For \textbf{RQ1},
we conclude that the change in the \emph{ECI} between clean and adversarial examples,
allows us to distinguish  between the actual contributed subgraph structures and adversarial perturbations.}
\end{framed}

\subsection{Performance of EGC\texorpdfstring{$^2$}{EGC\textasciicircum 2} on Clean Examples}
\label{clean_example}
{\color{black}In the previous section,
we distinguish between the actual contributed subgraph structures and adversarial perturbations on the graph by observing their differences in \emph{ECI}.
However,
an essential prerequisite for EGC$^2$ to improve the robustness of graph classification models is that it should ensure classification accuracy for clean examples.
To analyze the different effects of feature read-out and graph compression mechanisms on clean examples,
we conducted an ablation experiment.
Specifically,
we removed the read-out or graph compression mechanism in EGC$^2$,
and obtain two variants, Diff+Com and EGC.
Here,
we can answer question \textbf{RQ2} by comparing the performance of SAGPool, HGP-SL, Diffpool, EGC, Diff+Com, and EGC$^2$ on each clean dataset.}

\begin{table}[htbp]
	\centering
	\Huge
	\caption{Average graph classification accuracy$\pm$std on the clean graphs.
 We use bold to highlight wins.
 Methods with EGC indicate that the method has the feature read-out mechanism.}
	\resizebox{\textwidth}{!}{
	\begin{tabular}{cccccccccccc}
		\hline
		\multirow{2}{*}{Categories} & \multicolumn{1}{c}{\multirow{2}{*}{Baselines}} & \multicolumn{10}{c}{Accuracy$\pm$Std(\%)}                                                                                         \\ \cline{3-12}
		& \multicolumn{1}{c}{}                           &  PTC            &  PROTEINS       &  DD             &  NCI1           &  NCI109         &  IMDB-BINARY    &  REDDIT-BINARY &  OGBL-MOLHIV &  TWITCH\_EGOS &  FIRSTMM\_DB   \\ \hline
		\multirow{3}{*}{Kernel}     & SP                                              &  61.00$\pm$3.68          &  75.71$\pm$2.56          &  76.72$\pm$3.83          &  67.44$\pm$2.21          &  64.72$\pm$2.64          &  69.40$\pm$2.85          &  83.51$\pm$2.76  &85.12$\pm$2.42   & 63.21$\pm$2.94 & 38.79$\pm$4.87       \\
		& WL                                              &  67.52 $\pm$4.37         &  76.16$\pm$3.82          &  75.44$\pm$2.68          &  73.65$\pm$2.62          &  73.19$\pm$2.45          &  72.32$\pm$2.63          &  85.20$\pm$3.18   &  86.24$\pm$2.51  &  64.33$\pm$2.73  &  36.65$\pm$5.54        \\ 
		& WKPI                                              &  68.94$\pm$2.62 &  77.69$\pm$2.03 &  80.42$\pm$1.86 &  77.28$\pm$1.64 &  78.53$\pm$2.16 &  75.46$\pm$1.65 &  86.37$\pm$2.23 &  96.32$\pm$1.89 &  68.45$\pm$2.72 &  59.72$\pm$3.46        \\ \hline
		\multirow{3}{*}{GNN}        & GCN                                             &  68.75$\pm$2.51         &  74.54$\pm$3.39          &  77.17$\pm$2.36          &  72.90$\pm$1.87          &  72.14$\pm$1.85          &  72.31$\pm$1.33          &  84.17$\pm$2.48     &  96.03$\pm$1.21  &  67.51$\pm$2.61 &  41.67$\pm$3.79      \\
		& GAT                                             &  69.30$\pm$3.63          &  74.06$\pm$4.16          &  76.42$\pm$2.13          & 74.57$\pm$1.68          &  75.10$\pm$2.36          &  73.10$\pm$1.48          &  84.82$\pm$2.15       &  95.40$\pm$1.06 &  68.41$\pm$2.58 &  45.36$\pm$3.63   \\
		&  Ugformer &  70.53$\pm$2.26 &  79.36$\pm$1.72 &  80.68$\pm$1.57 &  76.30$\pm$1.62 &  76.38$\pm$1.68 &  75.69$\pm$0.84 &  86.81$\pm$1.43 &  96.68$\pm$0.86 &  68.59$\pm$1.92 & 60.76$\pm$2.38    \\ \hline
		\multirow{3}{*}{Pooling}    & Diffpool                                        &  68.00$\pm$2.21          &  76.25$\pm$2.95          &  80.91$\pm$1.21          &  74.29$\pm$1.53          &  74.10$\pm$ 1.49         &  75.20$\pm$1.06         &  86.19$\pm$1.43  &  96.13$\pm$0.85  &  68.49$\pm$2.31     &  57.14$\pm$3.26        \\
		& SAGPool                                         &  67.39$\pm$1.98         &  73.30$\pm$1.76        &  76.45$\pm$2.06          &  74.18$\pm$1.74          &  74.06$\pm$1.52         &  72.20$\pm$0.82         &  73.90$\pm$1.17   &  96.51$\pm$0.92   &  68.53$\pm$2.06  &  54.33$\pm$2.88         \\
		& HGP-SL                                          &  70.65$\pm$1.73         &  84.82$\pm$1.58          &  80.54$\pm$1.84          &  78.45$\pm$1.16          &  80.67$\pm$1.67          &  76.52$\pm$0.69          &  88.74$\pm$1.36  &  96.76$\pm$0.63   &  68.72$\pm$1.85   &  61.54$\pm$2.43         \\ \hline
		\multirow{5}{*}{Proposed}   & EGC$\rm_{GCN}$                                        &  72.65$\pm$2.38        &  76.36$\pm$2.65          &  82.81$\pm$1.73         &  68.21$\pm$1.69          &  68.98$\pm$1.53           &  75.25$\pm$1.26          &  87.12$\pm$2.24  &  96.39$\pm$1.13   &  69.92$\pm$1.52   &  51.33$\pm$2.97          \\
		& EGC$\rm _{GAT}$                                         &  74.52$\pm$3.41         &  75.59$\pm$2.84          &  82.03$\pm$1.68          &  70.57$\pm$1.14         &  72.05$\pm$1.82         &  74.10$\pm$1.41          &  86.21$\pm$1.76    &  96.82$\pm$0.84 &  70.23$\pm$1.38 &  54.42$\pm$2.42       \\
		& EGC$\rm _{Diffpool}$                                    &  \textbf{79.41$\pm$1.68}         &  83.63$\pm$2.53          &  82.73$\pm$1.05 &  \textbf{79.09$\pm$1.28}&  \textbf{80.92$\pm$1.26}&  \textbf{81.42$\pm$0.78} &  \textbf{94.55$\pm$1.35}    &  97.28$\pm$0.76   &  72.06$\pm$1.87  &  \textbf{66.67$\pm$2.71} \\
		& EGC$\rm _{SAGPool}$                                     &  73.53$\pm$1.54         &  79.09$\pm$1.67          &  80.91$\pm$1.33          &  75.45$\pm$0.89          &  76.71$\pm$1.13         &  78.60$\pm$0.84          &  80.00$\pm$0.89    &  97.09$\pm$0.87 &  71.64$\pm$0.96 &  62.25$\pm$2.26      \\
		& EGC$\rm _{HGP-SL}$                                      &  76.47$\pm$1.42&  \textbf{86.61$\pm$1.26} &  \textbf{84.03$\pm$1.52}          &  73.86$\pm$0.81         &  74.24$\pm$2.38         &  80.52$\pm$0.62         &  91.24$\pm$1.04   &  \textbf{97.64$\pm$0.68} &   \textbf{73.28$\pm$1.04} &   65.48$\pm$2.64       \\ \hline
    	\end{tabular}}
    	\label{tab:2}%
\end{table}

\begin{table}[htbp]
	\centering
	\Huge
	\caption{Average graph classification balanced accuracy$\pm$std on the clean graphs.
 We use bold to highlight wins.
 Methods with EGC indicate that the method has the feature read-out mechanism.}
	\resizebox{\textwidth}{!}{
		\begin{tabular}{cccccccccccc}
			\hline
			\multirow{2}{*}{Categories} & \multicolumn{1}{c}{\multirow{2}{*}{Baselines}} & \multicolumn{10}{c}{Balanced accuracy$\pm$Std(\%)}                                                                                         \\ \cline{3-12}
			& \multicolumn{1}{c}{}                          &   PTC            &   PROTEINS       &   DD            &   NCI1           &   NCI109        &   IMDB-BINARY    &   REDDIT-BINARY &   OGBL-MOLHIV &   TWITCH\_EGOS &   FIRSTMM\_DB   \\ \hline
			\multirow{3}{*}{Kernel}     &  SP &  60.84$\pm$3.75 &  73.46$\pm$2.65 &  74.61$\pm$3.68 &  68.52$\pm$2.16 &  64.58$\pm$2.61 &  70.28$\pm$2.93 &  82.34$\pm$2.86 &  79.67$\pm$2.73 &  63.87$\pm$3.16 & 34.67$\pm$4.78        \\
			&  WL &  67.24$\pm$4.13 &  73.82$\pm$3.62 &  74.18$\pm$2.47 &  71.89$\pm$2.53 &  73.64$\pm$2.48 &  71.46$\pm$2.57 &  84.67$\pm$3.02 &  81.54$\pm$2.56 &  64.39$\pm$2.85 & 32.48$\pm$5.16       \\ 
			& WKPI &  68.57$\pm$2.56 &  75.38$\pm$1.83 &  79.72$\pm$1.91 &  76.85$\pm$1.52 &  78.67$\pm$2.04 &  75.23$\pm$1.28 &  85.72$\pm$2.07 &  90.68$\pm$1.67 &  66.72$\pm$2.36 & 54.60$\pm$3.21    \\ \hline
			\multirow{3}{*}{ GNN}        &  GCN      &  68.63$\pm$2.44 &  74.15$\pm$3.44 &  77.35$\pm$2.43 &  72.67$\pm$1.74 &  71.86$\pm$1.82 &  71.15$\pm$1.04 &  82.25$\pm$2.13 &  87.62$\pm$1.83 &  66.92$\pm$2.61 &  43.72$\pm$3.84    \\
			&  GAT      &  68.72$\pm$3.26 &  73.78$\pm$4.03 &  75.82$\pm$2.18 &  73.85$\pm$1.48 &  74.79$\pm$2.16 &  73.92$\pm$1.46 &  83.59$\pm$2.08 &  89.27$\pm$1.58 &  68.76$\pm$2.43 &  47.38$\pm$3.52   \\
			 &  Ugformer &  68.95$\pm$1.73 &  76.42$\pm$1.96 &  78.86$\pm$1.48 &  76.53$\pm$1.65 &  76.14$\pm$1.52 &  74.73$\pm$0.81 &  86.57$\pm$1.46 &  91.28$\pm$0.73 &  68.97$\pm$1.79 &  58.64$\pm$2.65  \\ \hline
			\multirow{3}{*}{ Pooling}    &  Diffpool &  68.42$\pm$2.03 &  74.63$\pm$2.67 &  80.36$\pm$1.42 &  74.52$\pm$1.67 &  74.83$\pm$1.62 &  73.84$\pm$1.12 &  86.82$\pm$1.67 &  89.63$\pm$0.94 &  68.14$\pm$2.48 &  59.34$\pm$3.61       \\
			&  SAGPool  &  67.13$\pm$1.86 &  72.54$\pm$1.81 &  77.63$\pm$2.45 &  74.38$\pm$1.71 &  74.55$\pm$1.78 &  72.06$\pm$0.95 &  73.58$\pm$1.32 &  90.34$\pm$0.98 &  69.75$\pm$2.14 &  58.67$\pm$2.92     \\
			 &  HGP-SL   &  69.87$\pm$1.72 &  81.98$\pm$1.53 &  79.85$\pm$1.94 &  76.64$\pm$1.35 &  79.72$\pm$1.43 &  74.98$\pm$0.83 &  89.42$\pm$1.35 &  91.43$\pm$0.86 &  68.17$\pm$1.59 &  61.29$\pm$2.68         \\ \hline
			\multirow{5}{*}{ Proposed}   &  EGC$\rm_{GCN}$                                        &  73.67$\pm$2.64 &  76.48$\pm$2.47 &  82.36$\pm$1.65 &  70.36$\pm$1.62 &  69.74$\pm$1.75 &  74.18$\pm$1.14 &  86.59$\pm$2.16 &  89.67$\pm$1.28 &  69.47$\pm$1.52 &  58.64$\pm$2.73          \\
			&  EGC$\rm _{GAT}$                                         &  73.21$\pm$3.18 &  75.96$\pm$2.74 &  82.18$\pm$1.74 &  71.64$\pm$1.29 &  74.86$\pm$1.88 &  73.36$\pm$1.65 &  86.78$\pm$1.95 &  90.92$\pm$1.04 &  71.43$\pm$1.76 &  59.89$\pm$2.67      \\
			&  EGC$\rm _{Diffpool}$                                     &  \textbf{78.45$\pm$1.33} &  82.79$\pm$2.15          &  83.47$\pm$1.58          &  \textbf{78.98$\pm$1.36} &  \textbf{81.77$\pm$1.67} &  \textbf{80.73$\pm$0.91} &  \textbf{93.86$\pm$1.12} &  91.59$\pm$0.82          &  72.48$\pm$1.64          &  \textbf{70.21$\pm$2.54} \\
			&  EGC$\rm _{SAGPool}$                                     &  75.06$\pm$1.26          &  80.12$\pm$1.56          &  80.52$\pm$1.63          &  74.23$\pm$0.98          &  76.05$\pm$1.23          &  78.34$\pm$0.86          &  80.59$\pm$0.76          &  92.14$\pm$0.75          &  70.82$\pm$0.84          &  64.76$\pm$2.16      \\
			&  EGC$\rm _{HGP-SL}$                                     &  76.38$\pm$1.51          &  \textbf{84.12$\pm$1.35} &  \textbf{84.39$\pm$1.46} &  74.68$\pm$1.03          &  75.89$\pm$2.13          &  79.85$\pm$0.58          &  91.73$\pm$1.18          &  \textbf{92.76$\pm$0.62} &  \textbf{73.54$\pm$1.37} &  68.52$\pm$2.48      \\ \hline
	\end{tabular}}
	\label{tab:balanced}%
\end{table}

\subsubsection{Performance of features read-out}
\label{read-out}
{\color{black}First, we discuss the performance of EGC on clean examples.
The classification performance results are presented in TABLE \ref{tab:2}. The conclusions are as follows:}

\textbf{EGC achieves better classification performance.} TABLE \ref{tab:2} shows the classification performance of the proposed EGC family and all baselines.
{\color{black}The results indicated that the performance of the models with EGC was significantly better than that of the baseline models,
for instance,  the classification accuracy of HGP-SL and EGC$\rm_{HGP-SL}$ reached 70.65\% and 76.47\% for PTC,
respectively,
with an approximately 6\% improvement in accuracy.
Moreover,
the proposed EGC achieved SOTA performance among all datasets.
Furthermore,
considering the issue of unbalanced class distributions in datasets,
we adopted balanced accuracy to measure the performance of the graph classification methods in TABLE \ref{tab:balanced}.
A series variant of EGC reached a balanced accuracy of 78.00\% across ten datasets,
whereas kernel methods,
GNN and pooling achieved balanced accuracies of 70.68\%, 73.44\%, and 75.15\%
respectively.
This shows that a series of EGC variants can achieve the best classification performance among the baselines,
even considering the unbalanced distribution of the datasets.
They provided a positive answer to \textbf{RQ2}.
For instance,
EGC$\rm_{Diffpool}$ achieved the best classification accuracy of 80.92\%.
on the NCI109 dataset.}

{\color{black}\textbf{The feature read-out mechanism can improve  classification performance.} By comparing the GNN-based and graph pooling methods with the corresponding EGC methods,
we can conclude that the feature read-out mechanism can improve the classification performance of basic graph classification models in most cases.
Specifically,
for GNN-based methods and,
for instance, GCN and GAT,
the feature read-out mechanism in EGC, that is, EGC$\rm_{GCN}$ and EGC$\rm_{GAT}$,
improves the classification accuracy (average across ten datasets) by approximately 2.86\%.
Meanwhile,
the feature read-out mechanism can bring greater improvement (average 5.34\%) for hierarchical graph pooling methods as well,
that is, Diffpool, SAGPool, and HGP-SL,
across all ten datasets.}
This further indicates that as hierarchical graph pooling methods extract richer graph hierarchical information,
the feature read-out mechanism can aggregate this information into a more accurate global-level graph representation.

\textbf{The improvement brought by the feature read-out mechanism is related to the basic models and datasets.} {\color{black}As in the traditional methods,
graph kernel methods still exhibit considerable competitive performance, although they require immense human domain knowledge.
However,
we do not consider graph kernel methods as the basic model of EGC because they usually realize graph classification without a pooling operation.
For GNN-based and hierarchical graph-pooling methods,
the former obtains a minor performance improvement from the feature-read-out mechanism.
This is because the node-level representations they learned are still flat and
contains more low-value features.
In contrast,
hierarchical graph pooling methods learn the hierarchical structures of graphs,
which provides more valuable features for the feature-read-out mechanism.
Additionally,
We observed that for NCI1 and NCI109,
the original GNN-based methods and HGP-SL performed better.
We speculate that this may be because of the smaller scale and more straightforward structure of these datasets, resulting in the overfitting of the feature-read-out mechanism.
On the contrary,
the performance of the feature read-out mechanism is more stable on more complex datasets, such as PROTEINS, DD, and REDDIT-BINARY.}

\begin{figure}
	\centering
	\includegraphics[width=1 \linewidth]{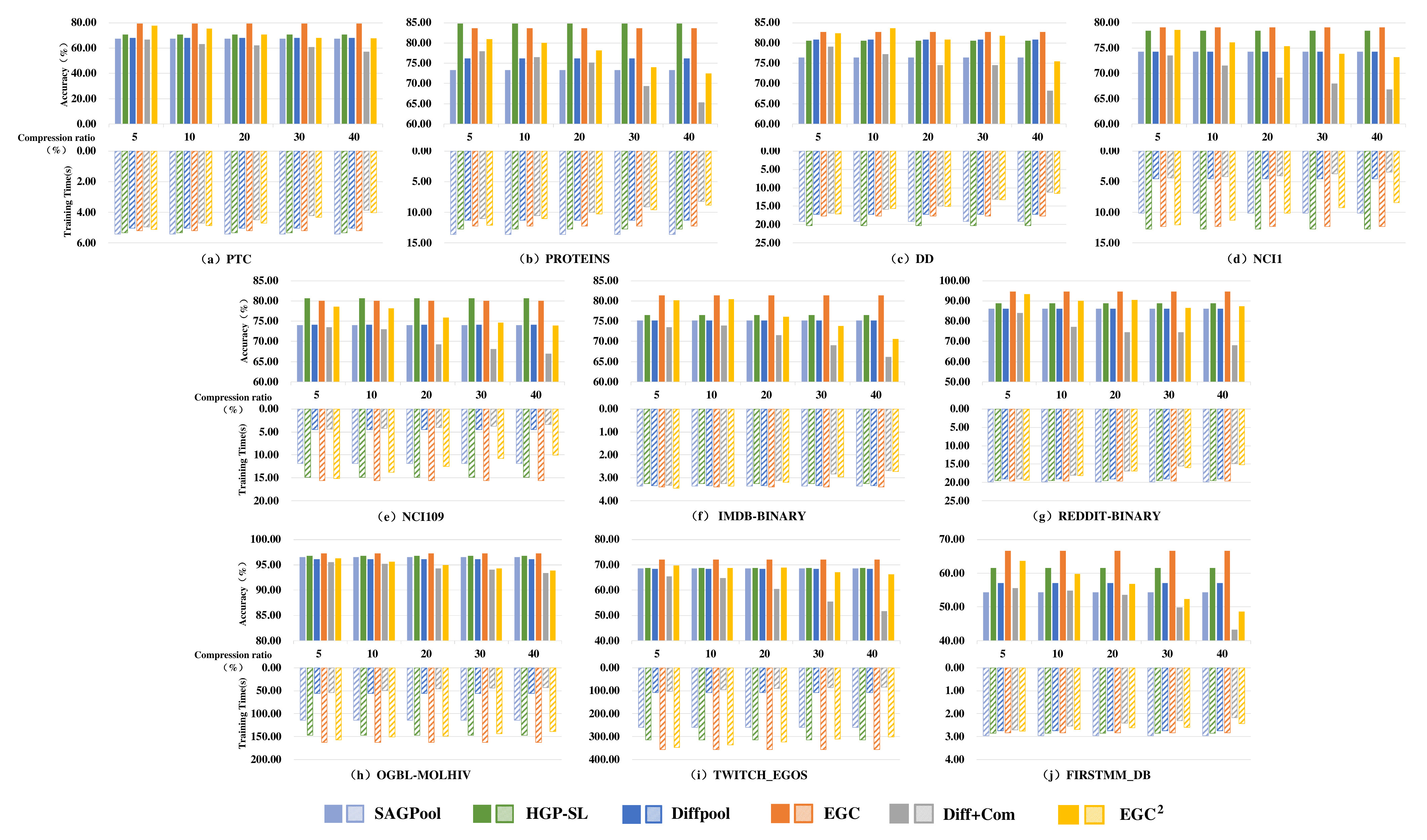}% 1\linewidth
	\caption{Ablation experiments with different compression ratios on clean examples.
 EGC is the EGC$^2$ without the graph compression mechanism.
 Diff+Com is the EGC$^2$ without the feature read-out mechanism.}\label{fig.7}
\end{figure}

\subsubsection{Performance of graph compression}
\label{sec5.7.2}
We further discuss the performance of EGC$^2$ on clean examples.
{\color{black}In this scenario,
we utilized C to preserve the contributed subgraph structures and explore the impact of different compression ratios $\gamma$.
Specifically,
for each graph,
we sorted the edges in descending order according to the C value and deleted lower-ordered edges and isolated nodes using the graph compression ratio $\gamma$.}
The values of $\gamma$ are 5\%, 10\%, 20\%, 30\%, and 40\%,
respectively.
{\color{black}Furthermore,
We use the SAGPool and HGP-SL, which perform the best under clean examples, as baselines to illustrate the effectiveness of graph compression.}
Fig. \ref{fig.7} shows the classification accuracy and the training time (per epoch) at the different $\gamma$ for each dataset.

{\color{black}First,
as $\gamma$ increased.
the classification accuracies of Diff+Com and EGC$^2$  decreased slightly.
When 40\% of the edges were removed,
the classification accuracy of Diff+Com and EGC$^2$ only decreased by 7.45\% and 5.88\%, respectively, compared with Diffpool and EGC on the NCI1 dataset,
respectively.
Even in this case,
EGC$^2$ can still achieve comparable or even better classification accuracy than Diffpool.

However,
the graph-compression mechanism can improve the training speed of the basic model.
Considering a compression ratio of $\gamma$ = 40\%,
the graph compression mechanism improves the EGC training speed by approximately 28.06\% on the PROTEINS dataset (the training seed EGC and EGC$^2$ are 12.26 and 8.82 s,
respectively).
Furthermore,
we noticed that the training speed improvement of Diff+Com,
that is, 11.43 s,
on the DD dataset was 33.74\% higher than that of Diffpool,
that is, 17.25 s,
which is better than the improvement on the other datasets when $\gamma$ = 40\%.
Compared with compressing other datasets,
a DD with a lower average node degree is more likely to produce isolated nodes,
thus reducing the scale of the graphs.
which is conducive to faster training of the model.}

\begin{framed}
{\color{black}Experiments demonstrated that the feature read-out mechanism can effectively improve the classification performance of graph classification models,
mainly when applied to hierarchical graph classification models.
Additionally,
the graph compression mechanism can accelerate the training speed without significantly decreasing the classification performance.
In summary,
for \textbf{RQ2},
EGC$^2$ can achieve comparable or even better classification accuracy than basic models.}
\end{framed}

\begin{figure}[htbp]
	\centering
	\includegraphics[width=1\linewidth]{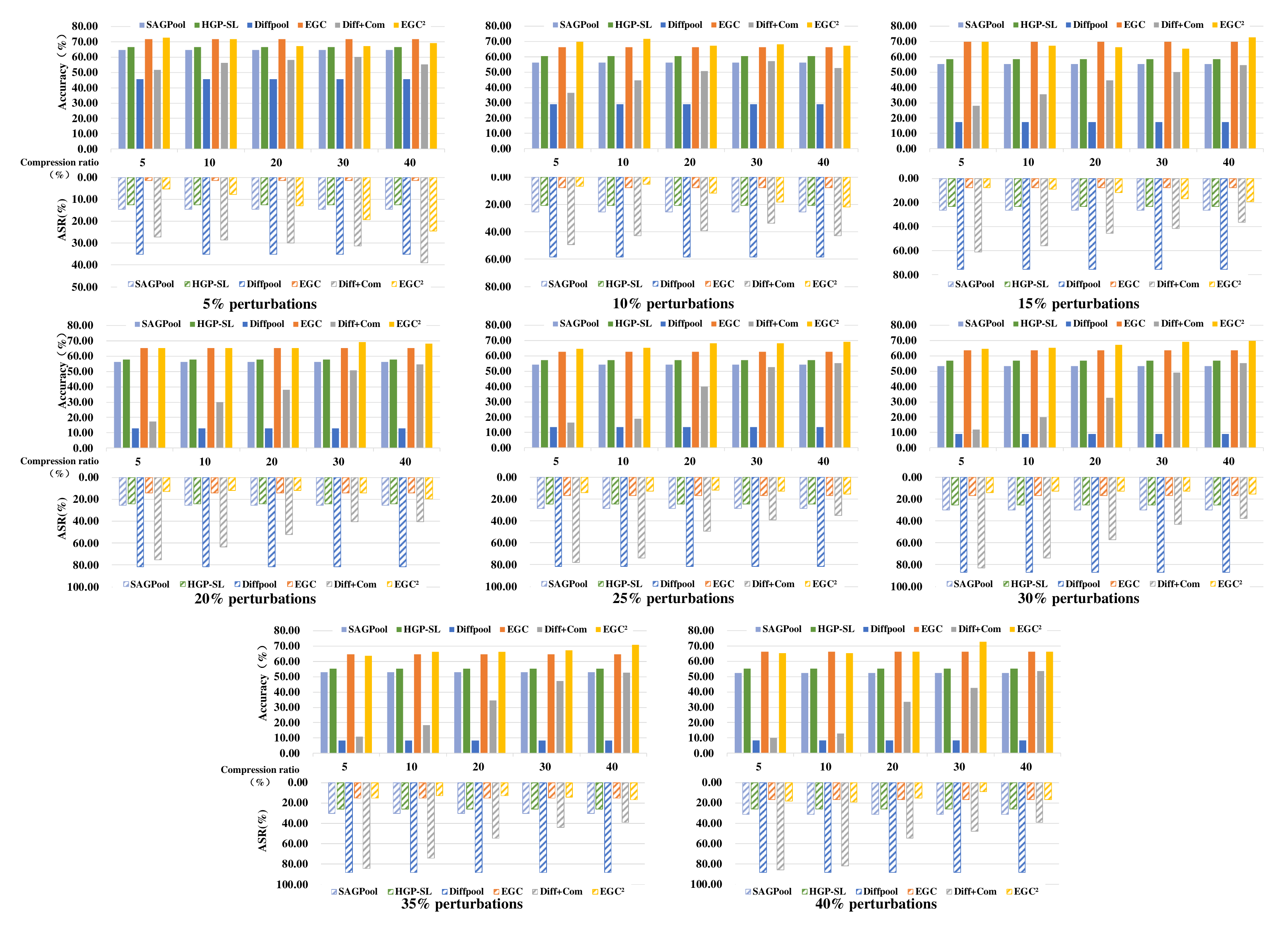}% 1\linewidth
	\caption{Ablation experiments of different compression ratios on adversarial examples with various perturbation ratios on PROTEINS.
 EGC is the EGC$^2$ without the graph compression mechanism. Diff+Com is the EGC$^2$ without the feature read-out mechanism.}\label{fig.8}
\end{figure}

\subsection{Defense Ability and Transferability of EGC \texorpdfstring{$^2$}{EGC\textasciicircum 2}}
\label{defense}
The previous section has ensured that the graph compression mechanism can effectively preserve the actual contributed subgraph structures on clean examples.
In this part,
we keep the same graph compression settings in section \ref{sec5.7.2}.
Then we use the adversarial attack methods introduced in section \ref{sec5.4} to generate adversarial examples and address questions \textbf{RQ3} and \textbf{RQ4}.

\begin{figure}
	\centering
	\includegraphics[width=0.8\linewidth]{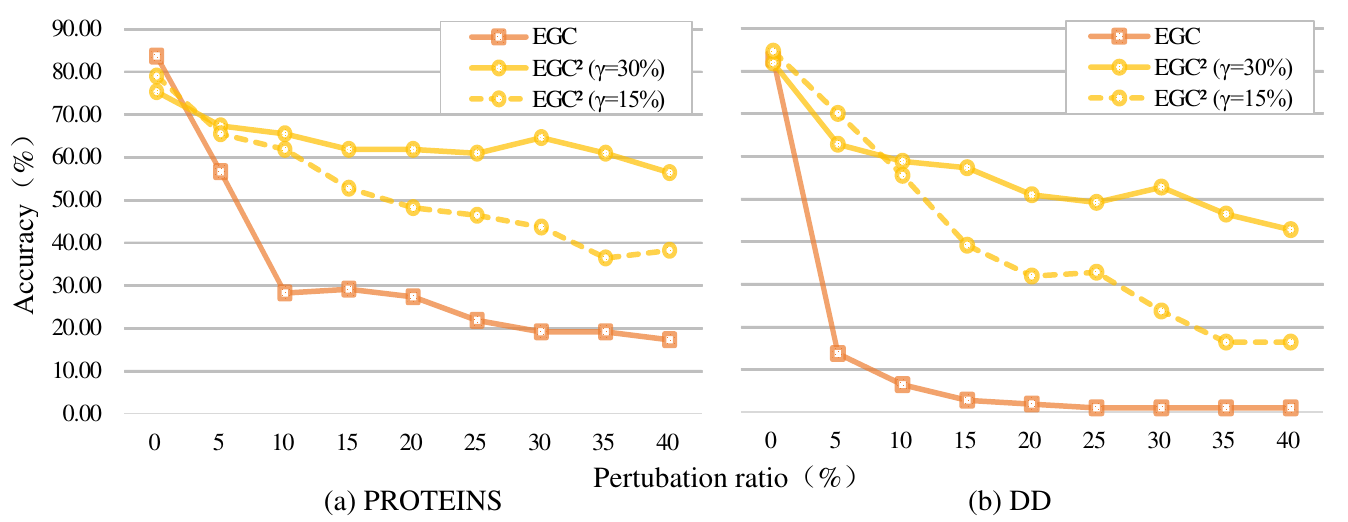}\\
	\caption{The defense ability of graph compression against adversarial attacks with different perturbation ratios in the compression ratio of 15\% and 30\%, respectively.
 EGC is the EGC$^2$ without the graph compression mechanism.} 
	\label{fig.9}
\end{figure}

\subsubsection{The defense ability on adversarial examples}
\label{5.8.1}
\textbf{Selection of compression ratio.} We first take Diffpool as the target model to generate adversarial examples by FGA,
and transfer the adversarial examples to other models.
Therefore,
under different perturbation attacks, the performance of the Diffpool model is the worst.
Fig. \ref{fig.8} shows the classification accuracy and the ASR with different graph compression ratios $\gamma$ on PROTEINS.

{\color{black}It is observed that as the perturbation ratio increases,
the classification accuracy on Diffpool drops sharply with the ASR rising rapidly.}
At the same time,
EGC can still achieve higher classification accuracy due to the additional feature read-out mechanism in Diffpool, 
although it is also affected by the adversarial perturbations.
Diff+Com also achieves much higher accuracy and lower ASR than  Diffpool.
The same phenomenon can also be observed between EGC$^2$ and EGC.
In addition,
the graph classification accuracy of the Diff+Com under a suitable compression ratio,
e.g., $\gamma$ = 30\%,
can be similar to that of SAGPool.
This verifies the effective defense ability of the graph compression mechanism.
For different compression ratios, with a small perturbation ratio (5\%-10\%), an overly large graph compression ratio ($>$20\%) will lead to an increase of ASR.
On the contrary,
as the perturbation ratio increases,
the higher graph compression ratios can achieve better classification accuracy with a decrease in ASR.
For instance,
in the face of 40\% perturbation,
the classification accuracy of the Diff+Com increases from 10.00\% to 53.64\% when the compression ratio increases from 5\% to 40\%.
Compared with the other five methods,
our EGC$^2$ can achieve the highest accuracy and the lowest ASR in most cases.

Considering that the perturbation ratio is usually not available in an actual scenario,
we select the $\gamma$ = 20\% (more stable) and $\gamma$ = 30\% (better defense ability) which are stable and have better defense ability in various perturbation ratios.

\textbf{Adaptive attack on EGC.} The adversarial attack with Diffpool as the target model can not be well applicable to SAGPool,
HGP-SL, EGC and EGC$^2$.
Fig.\ref{fig.9} illustrates the defense ability of the graph compression mechanism against the adaptive attacks on EGC.
When taking EGC as the target attack model,
the adaptive attack by FGA can also achieve satisfying attack effects,
which demonstrates that the feature read-out mechanism is insufficient under the adversarial attacks. In this scenario,
the graph compression mechanism can still effectively improve the robustness of EGC.
It is worth noting that at a smaller perturbation ratio,
the larger compression ratio ($\gamma$ = 30\%) performs worse,
since it may filter more actual contributed subgraph structures.
When the perturbation ratio increases, the $\gamma$ = 30\% performs better,
at this time,
the $\gamma$ = 15\% is not enough to reduce the impact of adversarial perturbations.

\begin{figure}
	\centering
	\includegraphics[width=1\linewidth]{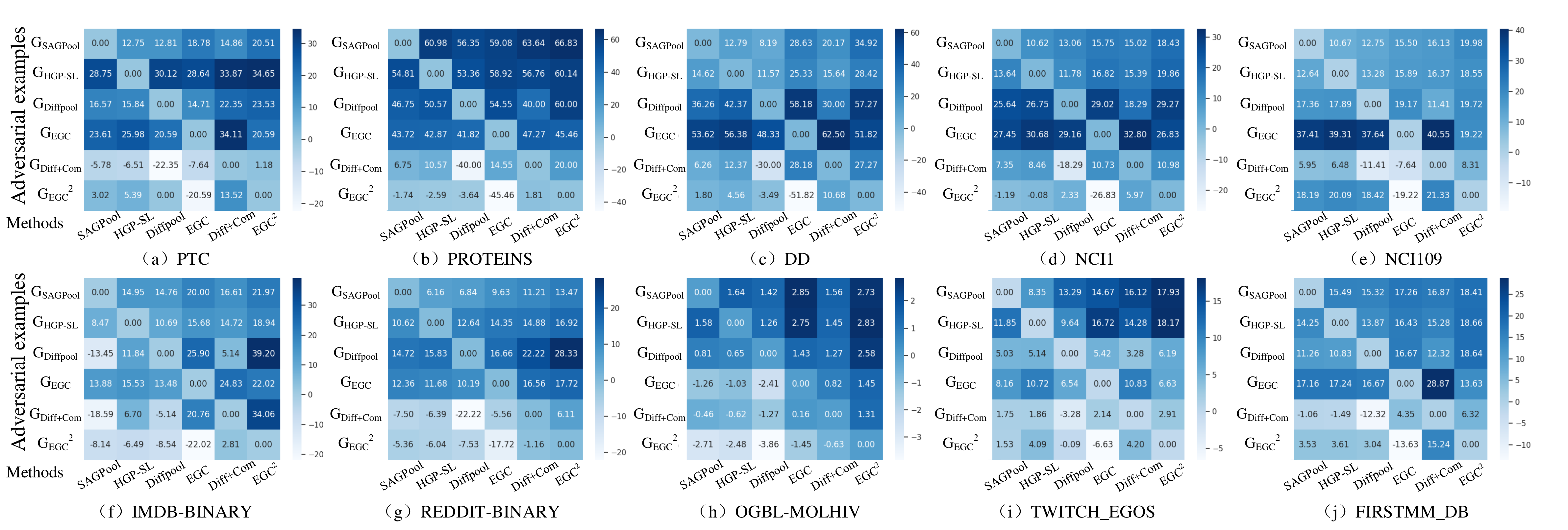}% 1\linewidth
	\caption{Average graph classification accuracy change heatmap under different adversarial examples.
 We take the accuracy on the diagonal as the benchmark value 0.
 For various adversarial examples,
 the graph compression mechanism can improve the classification accuracy of the basic model in most cases.
 EGC is the EGC$^2$ without the graph compression mechanism.
 Diff+Com is the EGC$^2$ without the feature read-out mechanism.}\label{fig.10}
\end{figure}

The performance of graph compression mechanism on other datasets when $\gamma$ = 30\% is shown in Fig.~\ref{fig.10}.
To illustrate the defense ability of the graph compression mechanism against different adversarial examples more intuitively,
we first generate adversarial examples based on six target models (including SAGPool, HGP-SL, Diffpool, EGC, Diff+Com, and EGC$^2$).
Since the graph compression mechanism is independent of model parameters,
the adversarial example caused by Diffpool is the same as the one caused by Diff+Com, and EGC and EGC$^2$ are similar.
Then,
we show the classification accuracy of the six models under different adversarial examples (with a perturbation ratio of 30\%,
which can reach almost the highest ASR).
We take the accuracy of different adversarial examples on the corresponding graph classification model as the reference value (0.00),
and give the decrease or increase value relative to the reference value in other cases.
{\color{black}We can see that,
the adversarial examples generated with SAGPool, HGP-SL,
Diffpool or EGC as the target model all have weak attack transferability.}
Therefore,
we pay more attention to improving the accuracy of the basic model brought by the graph compression mechanism.
Under different adversarial examples,
EGC$^2$ can achieve the highest classification accuracy in most cases.
In addition,
EGC$^2$ and Diff+Com show higher classification accuracy than EGC and Diffpool,
respectively,
indicating that the graph compression mechanism indeed enhances the robustness of the basic models.

\subsubsection{The transferability of EGC \texorpdfstring{$^2$}{EGC\textasciicircum 2}}
\label{5.8.2}
In the previous experiments,
we regard Diffpool as the basic model of EGC$^2$,
and explore the defense performance of the graph compression mechanism through adversarial examples generated by FGA.
Next,
we will illustrate the transferability of EGC$^2$ from two different scenarios:
1) the defense performance for different basic models,
and 2) the defense performance under different attack methods.

\begin{table}[htbp]
	\centering
	\Huge
	\caption{The ASR of EGC$^2$ and its variants are based on different basic models under different attack methods, the lower ASR is better.
 We use bold to highlight wins. Methods with EGC indicates that the method has the feature read-out mechanism.
 Methods with Com indicates that the method has the graph compression mechanism.
 Values in $\left(\right)$ denote the percentage of ASR reduction for the method compared to its baseline method.}
	\resizebox{\textwidth}{!}{\begin{tabular}{ccccccccccccc}
		\hline
		\multirow{2}{*}{\begin{tabular}[c]{@{}c@{}}Attack \\ method\end{tabular}} & \multirow{2}{*}{\begin{tabular}[c]{@{}c@{}}Target  \\ model\end{tabular}} & \multirow{2}{*}{Method} & \multicolumn{10}{c}{ASR(\%)}                                                                                                                                          \\ \cline{4-13}
		&                                &                         & PTC                   & PROTEINS              & DD                    & NCI1                  & NCI109               & IMDB-BINARY           & REDDIT-BINARY &   OGBL-MOLHIV &   TWITCH\_EGOS &   FIRSTMM\_DB        \\ \hline
		\multirow{12}{*}{FGA}          & \multirow{2}{*}{EGC$\rm_{GCN}$ }       & EGC$\rm_{GCN}$                 & 75.71                 & 88.10                 & 98.90                 & 65.36                 & 65.73                & 54.58                 & 32.16 &   3.66 &  96.96 &  43.68                 \\
		&                                & EGC$^2\rm_{GCN}$                 & \textbf{36.35 (51.98$\downarrow$)} & \textbf{46.82 (46.86$\downarrow$)} & \textbf{68.00 (31.21$\downarrow$)}  & \textbf{29.37 (55.07$\downarrow$)} & \textbf{42.80 (34.78$\downarrow$)} & \textbf{33.45 (38.71$\downarrow$)} & \textbf{7.13 (77.84$\downarrow$)} &   \textbf{2.62 (28.42$\downarrow$)}&   \textbf{67.78 (30.09$\downarrow$)}&   \textbf{26.42 (39.51$\downarrow$)} \\ \cline{2-13}
		& \multirow{4}{*}{GCN}           & GCN                     & 78.60                 & 89.03                 & 98.82                 & 90.03                 & 81.09                & 83.21                 & 84.88  &  8.27         &  88.65 &  68.33       \\
		&                                & GCN+Com                 & 48.67 (38.08$\downarrow$)          & 51.22 (42.47$\downarrow$)          & 64.20 (34.97$\downarrow$)           & 57.60 (36.02$\downarrow$)          & 59.60 (26.41$\downarrow$)          & 49.19 (40.88$\downarrow$)          & 47.07 (44.54$\downarrow$) &   6.14 (25.76$\downarrow$) &   60.49 (31.77$\downarrow$)&   56.75 (16.95$\downarrow$)           \\
		&                                & EGC$\rm_{GCN}$                  & 57.22 (27.20$\downarrow$)          & 78.05 (12.33$\downarrow$)          & 86.30 (12.63$\downarrow$)           & 56.35 (37.41$\downarrow$)          & 48.30 (40.39$\downarrow$)         & 34.14 (58.97$\downarrow$)          & 4.71 (28.69$\downarrow$)  &   5.71 (30.96$\downarrow$) &   53.37 (39.80$\downarrow$) &   36.67 (46.33$\downarrow$)         \\
		&                                & EGC$^2\rm_{GCN}$                 & \textbf{39.94 (49.19$\downarrow$)} & \textbf{45.12 (49.32$\downarrow$)} & \textbf{49.10 (50.30$\downarrow$)}  & \textbf{36.41 (59.56$\downarrow$)} & \textbf{45.30 (44.05$\downarrow$)} & \textbf{3.47 (95.83$\downarrow$)}  & \textbf{3.87 (95.44$\downarrow$)} &   \textbf{4.09 (50.54$\downarrow$)} &   \textbf{48.94 (44.79$\downarrow$)} &   \textbf{32.53 (52.38$\downarrow$)} \\ \cline{2-13}
		& \multirow{2}{*}{EGC$\rm_{SAG}$ }       & EGC$\rm_{SAG}$                  & 60.00                 & 50.09                 & 73.25                 & 56.62                 & 52.60                & 47.15          & 25.95        &  7.62   &  93.19 &  46.83               \\
		&                                & EGC$^2\rm_{SAG}$                 & \textbf{36.41 (39.32$\downarrow$)} & \textbf{40.35 (19.46$\downarrow$)}  & \textbf{67.64 (7.66$\downarrow$)}  & \textbf{32.52 (42.56$\downarrow$)} & \textbf{42.88 (18.49$\downarrow$)} & \textbf{31.62 (32.95$\downarrow$)} & \textbf{15.99 (38.39$\downarrow$)}  &   \textbf{4.19 (45.01$\downarrow$)}  &   \textbf{44.80 (51.93$\downarrow$)}  &   \textbf{41.29 (11.83$\downarrow$)}  \\ \cline{2-13}
		& \multirow{4}{*}{SAGPool}           & SAGPool                     & 47.63                 & 50.27                 & 79.79                 & 80.39                 & 32.49                & 34.81                 & 28.65    &  5.76 &  87.54 &  64.58              \\
		&                                & SAG+Com                 & 43.27 (9.16$\downarrow$)           & 33.03 (34.30$\downarrow$)          & 75.13 (5.84$\downarrow$)           & 46.08 (42.68$\downarrow$)          & 29.91 (7.94$\downarrow$)          & 29.72 (14.60$\downarrow$)           & \textbf{13.15 (54.09$\downarrow$) }    &   4.71 (18.23$\downarrow$)&   35.56 (59.38$\downarrow$) &  58.16 (9.94$\downarrow$)     \\
		&                                & EGC$\rm_{SAG}$                  & 25.81 (45.83$\downarrow$)          & 28.06 (44.18$\downarrow$)          & 50.06 (37.26$\downarrow$)           & 27.69 (65.55$\downarrow$)          & \textbf{21.44 (34.00$\downarrow$)} & 25.26 (27.42$\downarrow$)           & 17.58 (38.64$\downarrow$)   &   3.66 (36.46$\downarrow$)  &   35.48 (59.47$\downarrow$)  &   46.32 (28.28$\downarrow$)         \\
		&                                & EGC$^2\rm_{SAG}$                 & \textbf{17.08 (64.14$\downarrow$)} & \textbf{21.87 (56.50$\downarrow$)} & \textbf{47.50 (40.39$\downarrow$)}  & \textbf{17.89 (77.75$\downarrow$)} & 27.59 (15.09$\downarrow$)          & \textbf{21.50 (38.24$\downarrow$)} & \textbf{13.15 (54.09$\downarrow$)} &   \textbf{3.14 (45.49$\downarrow$)}&   \textbf{20.79 (76.25$\downarrow$)}&   \textbf{42.05 (34.89$\downarrow$)} \\ \hline
		\multirow{6}{*}{PoolAttack}    & \multirow{2}{*}{EGC$\rm_{SAG}$ }       & EGC$\rm_{SAG}$                  & 16.01                 & 5.75                  & 15.40                 & \textbf{13.25}        & 30.07                & 16.28                 & 8.10    &  4.58 &  68.52 &  38.64             \\
		&                                & EGC$^2\rm_{SAG}$                 & \textbf{4.00 (75.02$\downarrow$)}  & \textbf{4.60 (20.00$\downarrow$)}   & \textbf{10.23 (33.55$\downarrow$)}  & 24.10 (81.80$\uparrow$)          & \textbf{23.80 (20.85$\downarrow$)} & \textbf{13.47 (17.27$\downarrow$)}  & \textbf{4.76 (41.20$\downarrow$)}  &   \textbf{2.46 (46.29$\downarrow$)}&   \textbf{43.93 (35.89$\downarrow$)}&   \textbf{33.72 (12.73$\downarrow$)}  \\ \cline{2-13}
		& \multirow{4}{*}{SAGPool}           & SAGPool                     & 30.17                 & 10.08                 & 24.00                 & 21.57                 & 18.98                & 17.63                 & 2.61  &  3.64 &  57.86 &  58.42                \\
		&                                & SAG+Com                 & 21.44 (28.92$\downarrow$)           & 3.26 (67.66$\downarrow$)            & 17.41 (27.47$\downarrow$)           & 24.63 (14.19$\uparrow$)          & \textbf{14.81 (21.98$\downarrow$)} & \textbf{9.72 (44.85$\downarrow$)}   & 1.01 (61.14$\downarrow$)    &   4.31(18.41$\uparrow$)    &   49.74 (14.03$\downarrow$)   &  50.03 (14.36$\downarrow$)           \\
		&                                & EGC$\rm_{SAG}$                 & \textbf{8.35 (72.31$\downarrow$)}           & 7.87 (21.92$\downarrow$)            & 15.34 (36.08$\downarrow$)           & \textbf{20.34 (5.69$\downarrow$)}  & 18.98 (0.00$\downarrow$)          & 12.02 (31.81$\downarrow$)           & 1.19 (54.40$\downarrow$)      &   2.83 (22.25$\downarrow$)      &   50.28 (13.10$\downarrow$)      &   36.58 (37.38$\downarrow$)           \\
		&                                & EGC$^2\rm_{SAG}$                & \textbf{8.35 (72.31$\downarrow$)}  & \textbf{2.02 (79.97$\downarrow$)}   & \textbf{12.62 (47.41$\downarrow$)} & \textbf{20.34 (5.69$\downarrow$)}  & 16.61 (12.52$\downarrow$)          & 11.58 (34.33$\downarrow$)           & \textbf{0.00 (100.00$\downarrow$)}  &   \textbf{2.26 (37.91$\downarrow$)}&   \textbf{38.64 (33.22$\downarrow$)}&   \textbf{32.64 (44.13$\downarrow$)}  \\ \hline
	\end{tabular}}
	\label{tab3}
\end{table}

\textbf{EGC$^2$ with multiple basic models.}
For the first scenario,
we choose the GNN-based method GCN and the node-sampling-based graph pooling method SAGPool as the basic model of EGC$^2$,
respectively. Here we still generate adversarial examples through FGA with a perturbation ratio of 30\%.
Also, we only retain the compression ratio of $\gamma$ = 30\% since it can achieve better defense performance in most cases.
The ASR (and the relative percentage change in ASR to the basic model) of EGC$^2$ and its variants based on different basic models under different attacks are shown in TABLE \ref{tab3}.
We can conclude that the defense ability of the graph compression mechanism is independent of the basic models.
Taking GCN as the attack target of FGA,
the GCN+Com, EGC$\rm_{GCN}$ and EGC$^2\rm_{GCN}$ decrease the ASR  by 38.08\%, 27.20\% and 49.19\% on PTC dataset,
respectively.
Due to the node sampling pooling operation of SAGPool,
it has a more difficult classification boundary  to cross than GCN and Diffpool,
thus the improvement brought by the graph compression mechanism may be slightly affected,
e.g., the ASR of SAG+Com, EGC$\rm_{SAG}$ and EGC$^2\rm_{SAG}$ decreased by 14.60\%, 27.42\% and 38.24\% compared with SAGPool on IMDB-BINARY dataset, respectively. 

\textbf{Adversarial graphs generated by PoolAttack. }For the second scenario,
PoolAttack is chosen to generate the adversarial examples.
Considering that it is only applicable to the graph pooling method based on node sampling,
here we take SAGPool as the basic model of EGC$^2$,
and maintain the same perturbation and graph compression ratios as before.
TABLE \ref{tab3} also indicates that,
as an indirect attack method targeting reserved nodes,
PoolAttack has a lower ASR on SAGPool than FGA.
In this scenario,
we can still observe the defense ability of the graph compression mechanism to against PoolAttack.

\begin{framed}
In this section,
we answer questions \textbf{RQ3} and \textbf{RQ4}. For \textbf{RQ3},
experiments demonstrate that EGC$^2$ has stronger robustness than the basic models at an appropriate compression ratio.
For \textbf{RQ4}, EGC$^2$ still has satisfying defense ability under different basic models or attack methods.
\end{framed}

\subsection{Visualization and Parameter Sensitivity Analysis  }
\label{sec.5.9}

\begin{figure}
	\centering
	\includegraphics[width=0.9\linewidth]{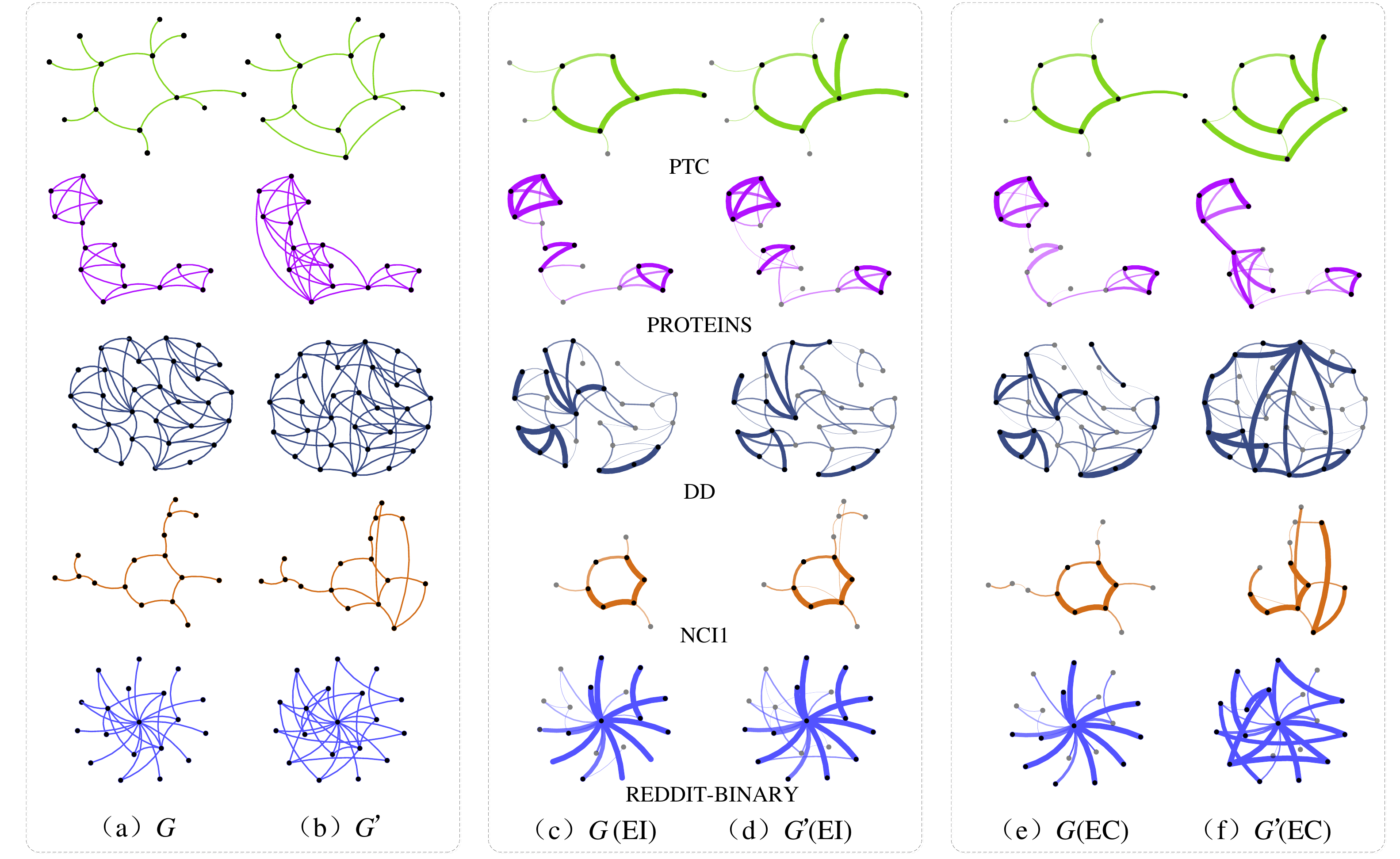}% 1\linewidth
	\caption{The visualization results of graph compression for different datasets.
 The thickness of the line represents different edge contribution/importance.
 (a) and (b) are original and adversarial examples without graph compression, respectively.
 (c) and (d) are compressed by edge importance index, and (e) and (f) are compressed by edge contribution.
 Graph compression by edge importance can better filter adversarial perturbations. }\label{fig.11}
\end{figure}

\textbf{Visualization.}
We utilize gephi to visualize graph compression results.
In detail,
we randomly select small-scale graphs from the five datasets of PTC, PROTEINS, DD, NCI1, REDDIT-BINARY.
We plot the original and the adversarial examples without graph compression,
and the compression examples which are compressed by the C and the edge gradient in Fig.~\ref{fig.11}.
Since the gradient information directly determines the classification result of the graph,
the compression examples in Fig.~\ref{fig.11}(e) preserve the actual contributed subgraph structures.
As we can see,
from the original examples,
both graph compression methods can preserve the actual contributed subgraph structures.
For the adversarial examples,
graph compression by edge importance can better filter adversarial perturbations.

\begin{figure}
% \setlength{\abovecaptionskip}{0.cm}
% 	\setlength{\belowcaptionskip}{-0.cm}
% \vspace{-1cm}
	\centering
	\includegraphics[width=0.8\linewidth]{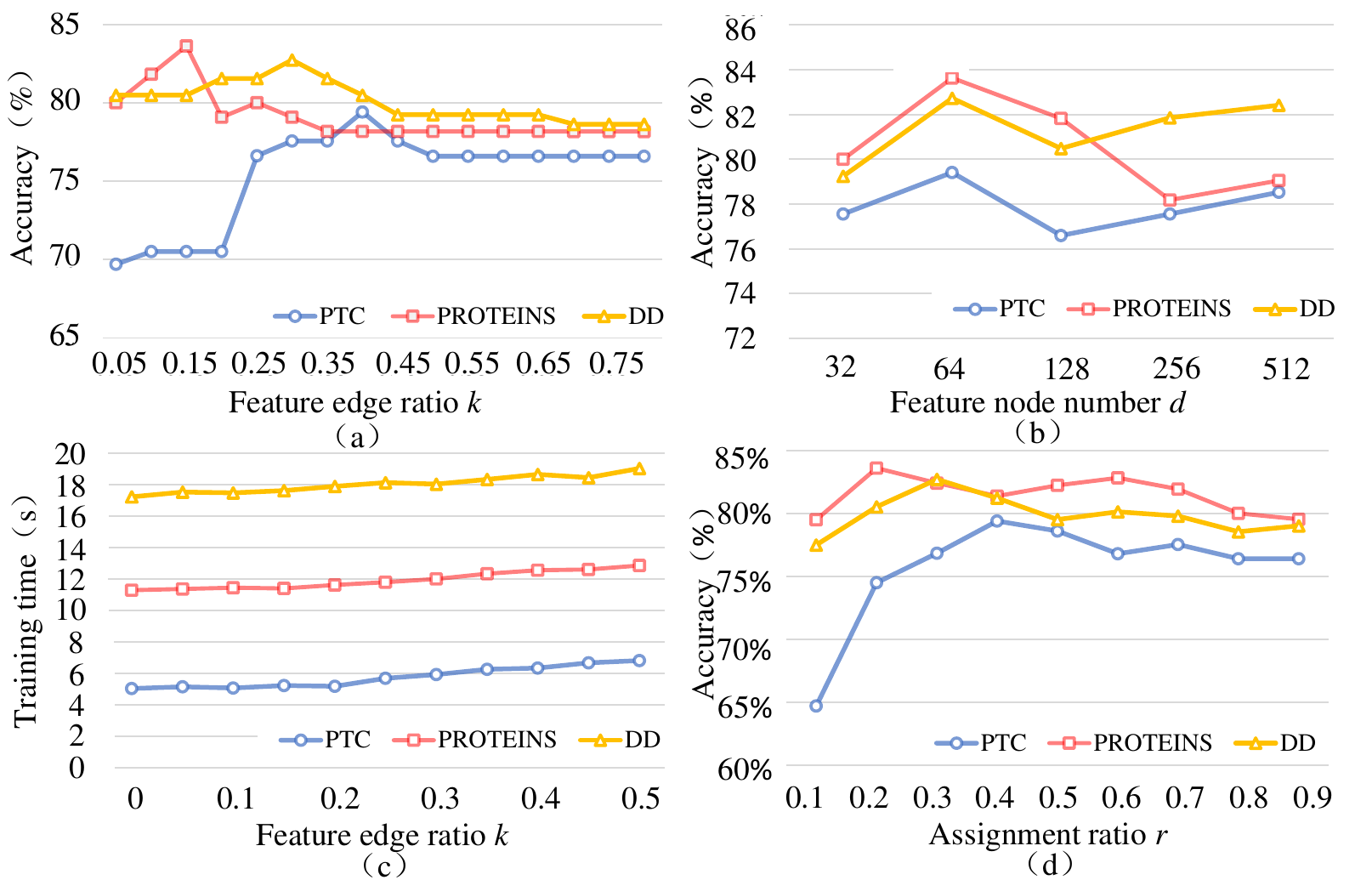}\\
	\caption{Hyper-parameter sensitivity analysis.}
	\label{fig.4}
\end{figure}

\textbf{Parameter sensitivity analysis.} We further study the sensitivity of several key hyper-parameters at different scales.
Specifically,
we take clean PTC, PROTEINS and DD as examples to illustrate the influence of the feature edge ratio $k$,
the number of feature nodes $d$,
and the assignment ratio $r$ of Diffpool on the performance of EGC$^2$.
As shown in Fig. \ref{fig.4} (a) and Fig. \ref{fig.4} (d), the best performance $k$ and $r$ are uncertain,
but in general, the best $k$ and $r$ are always less than 0.5. In Fig. \ref{fig.4} (b), EGC$^2$  almost achieves the best performance across different datasets when setting $d=64$.
Combined with Fig. \ref{fig.4} (b) and (c),
it can be demonstrated that the feature read-out mechanism does not lead to a noticeable increase in the time complexity of EGC$^2$.

\section{Conclusion}
{\color{black}In this study,
we propose an enhanced graph-classification model with easy graph compression,
called EGC$^2$. EGC$^2$ compensates for the performance degradation caused by the graph compression mechanism through the feature read-out mechanism,
thus achieving efficient and robust graph-classification performance.
The experimental results demonstrate SOTA performance of our method on ten benchmark datasets compared with other baseline graph classification methods.
Although the feature-read-out mechanism introduces additional parameters,
it does not cause a noticeable increase in complexity because the parameter dimension is controllable.
Moreover,
we designed an \emph{ECI} to distinguish between the actual contributed subgraph structures and adversarial perturbations,
which can help the graph compression mechanism preserve the actual contributed subgraph structures, filter trivial structures, and even adversarial perturbations.
Experiments on multiple basic models and adversarial attacks also demonstrated that EGC$^2$ has excellent transferability.

Robustness is a crucial factor in graph classification.
For future work,
We aim to explore a defense method with low complexity that avoids performance degradation.
Furthermore,
improving the robustness of the graph classification model under the threat of poisoning attacks warrants further research.}

%\section*{References}

%\newpage

%\appendix

\end{document}